\documentclass[nohyperref,accepted]{article}

\usepackage{microtype}
\usepackage{graphicx}
\usepackage{subcaption}
\usepackage{booktabs} 

\usepackage{hyperref}

\usepackage{icml2023}

\usepackage{amsmath}
\usepackage{amssymb}
\usepackage{mathtools}
\usepackage{amsthm}

\usepackage[capitalize,noabbrev]{cleveref}

\theoremstyle{plain}

\theoremstyle{definition}

\theoremstyle{remark}

\usepackage[textsize=tiny]{todonotes}

\usepackage{multirow}
\usepackage{ulem}
\usepackage{bm}
\usepackage{arydshln}

\newcommand{\std}[1]{\scalebox{0.6}{$\pm#1$}}

\icmltitlerunning{Parameter-Level Soft-Masking for Continual Learning}

\begin{document}

\twocolumn[
\icmltitle{Parameter-Level Soft-Masking for Continual Learning}



\icmlsetsymbol{visit}{\textdagger}

\begin{icmlauthorlist}
\icmlauthor{Tatsuya Konishi}{kddir,visit}
\icmlauthor{Mori Kurokawa}{kddir}
\icmlauthor{Chihiro Ono}{kddir}
\icmlauthor{Zixuan Ke}{uic}
\icmlauthor{Gyuhak Kim}{uic}
\icmlauthor{Bing Liu}{uic}
\end{icmlauthorlist}

\icmlaffiliation{kddir}{KDDI Research, Inc., Fujimino, Japan.}
\icmlaffiliation{uic}{University of Illinois at Chicago, Chicago, United States}

\icmlcorrespondingauthor{Tatsuya Konishi}{tt-konishi@kddi.com}

\icmlkeywords{Machine Learning, ICML}

\vskip 0.3in
]



\printAffiliationsAndNotice{
\textsuperscript{\textdagger}The work was done when this author was visiting Bing Liu's group at University of Illinois at Chicago.
}


\begin{abstract}
Existing research on \textit{task incremental learning} in continual learning has primarily focused on preventing \textit{catastrophic forgetting} (CF).
Although several techniques have achieved learning with no CF, they attain it by letting each task monopolize a sub-network in a shared network, which seriously limits knowledge transfer (KT) and causes over-consumption of the network capacity, i.e., as more tasks are learned, the performance deteriorates. 
The goal of this paper is threefold: (1) overcoming CF, (2) encouraging KT, and (3) tackling the capacity problem. 
A novel technique (called SPG) is proposed that \textit{soft-masks} (partially blocks) parameter updating in training based on the importance of each parameter to old tasks. 
Each task still uses the full network, i.e., no monopoly of any part of the network by any task, which enables maximum KT and reduction in capacity usage.
To our knowledge, this is the first work that soft-masks a model at the parameter-level for continual learning.
Extensive experiments demonstrate the effectiveness of SPG in achieving all three objectives. 
More notably, it attains significant transfer of knowledge not only among similar tasks (with shared knowledge) but also among dissimilar tasks (with little shared knowledge) while mitigating CF.
\end{abstract}

\section{Introduction} \label{sec:introduction}
Catastrophic forgetting (CF) and knowledge transfer (KT) are two key challenges of continual learning (CL), which learns a sequence of tasks incrementally. CF refers to the phenomenon where a model loses some of its performance on previous tasks once it learns a new task. KT means that tasks may help each other to learn by sharing knowledge.
This work further investigates these problems in the popular CL paradigm, \textit{task-incremental learning} (TIL). In TIL, each task consists of several classes of objects to be learned. Once a task is learned, its data is discarded and will not be available for later use. 
During testing, the task id is provided for each test sample so that the corresponding classification head of the task can be used for prediction.

Several effective approaches have been proposed for TIL that can achieve learning with little or no CF. \textit{Parameter isolation} is perhaps the most successful one in which the system learns to mask a sub-network for each task in a shared network.
HAT~\citep{hat} and SupSup~\citep{supsup} are two representative systems.
HAT set binary/hard masks on neurons (not parameters) that are important for each task. In learning a new task, those masks block the gradient flow through the masked neurons in the backward pass. Only those \textit{free} (unmasked) \textit{neurons} and their parameters are trainable. Thus, as more tasks are learned, the number of free neurons left becomes fewer, making later tasks harder to learn, which results in gradual performance deterioration (see \cref{sec:capacity}). Further, if a neuron is masked, all the parameters feeding to it are also masked, which consumes \textbf{a great deal of network capacity} (hereafter referred to as the ``capacity problem''). 
As the sub-networks for old tasks cannot be updated, it has \textbf{limited knowledge transfer}.  
CAT~\citep{cat} tries to improve KT of HAT by detecting task similarities. If the new task is found similar to some previous tasks, these tasks' masks are removed so that the new task training can update the parameters of these tasks for backward pass. However, this is risky because if a dissimilar task is detected as similar, serious CF occurs, and if similar tasks are detected as dissimilar, its knowledge transfer will be limited.
SupSup uses a backbone network (randomly initialized) and finds a sub-network for each task. The sub-network is represented by a mask, which is a set of binary gates indicating which parameters in the network are used. The mask for each task is saved. Since the network is not changed, SupSup has no CF or capacity problem, but since each mask is independent of other masks, SupSup by design has no KT.

To tackle these problems, we propose a very different approach, named ``\underline{\textit{S}}oft-masking of \underline{\textit{P}}arameter-level \underline{\textit{G}}radient flow'' (SPG). It is surprisingly effective and contributes in following ways:

(1).~Instead of learning hard/binary masks on neurons for each task and blocking these neurons in training a new task and in testing like HAT, SPG computes an importance score for each network parameter (not neuron) to old tasks using gradients. The reason that gradients can be used as importance is because gradients directly tell how a change to a specific parameter will affect the output classification and may cause CF. SPG uses the importance score of each parameter as a \textit{soft-mask} to constrain the gradient flow in the backward pass to ensure those important parameters to old tasks have minimum changes in learning a new task to prevent CF of previous knowledge. 
To our knowledge, soft-masking of parameters has not been done before. 

(2).~SPG has some resemblance to the popular regularization-based approach, e.g., EWC~\citep{ewc}, in that both use importance of parameters to constrain changes to important parameters of old tasks. But there is a major difference. SPG directly controls each parameter (fine-grained), but EWC controls all parameters together using a regularization term in the loss to penalize the sum of changes to all parameters in the network (rather coarse-grained). \cref{sec:results} shows that our \textbf{soft-masking is markedly better than regularization}. We believe this is an important result.

(3).~In the forward pass, no masks are applied, which encourages knowledge transfer among tasks. This is better than CAT as SPG does not need extra mechanism for task similarity comparison. Knowledge sharing and transfer in SPG are automatic. SupSup cannot do knowledge transfer.

(4).~As SPG soft-masks parameters, it does not monopolize any parameters or sub-network like HAT for each task and SPG's forward pass does not use any masks. This reduces the \textit{capacity problem}.

Experiments with the standard CL setup have been conducted with (1) similar tasks to demonstrate SPG's better knowledge transfer 
and (2) dissimilar tasks to show SPG's ability to overcome CF, and (3) deal with the capacity issue. None of the baselines is able to achieve all.
The code is available at \url{https://github.com/UIC-Liu-Lab/spg}.

\section{Related Work} \label{sec:related_work}
Approaches in continual learning can be grouped into three main categories. We review them below.

\textbf{Regularization-based}:
This approach computes importance values of either parameters or their gradients on previous tasks, and adds a regularization in the loss to restrict changes to those important parameters to mitigate CF. 
EWC~\citep{ewc} uses the Fisher information matrix to represent the importance of parameters and a regularization to penalize the sum of changes to all parameters. 
SI~\citep{si} extends EWC to reduce the complexity in computing the penalty.
Many other approaches~\citep{lwf,dmc,ucl} in this category have also been proposed, but they still have difficulty to prevent CF.  
As discussed in the introduction section, the proposed approach SPG has some resemblance to a  regularization based method EWC. But the coarse-grained approach of using regularization is significant poorer than the fine-grained soft-masking in SPG for overcoming CF as we will see in \cref{sec:results}.

\textbf{Memory-based}: 
This approach introduces a small memory buffer to store data of previous tasks and replay them in learning a new task to prevent CF~\citep{gem,a-gem}. Some methods~\citep{gr,binplay} prepare data generators for previous tasks, and the generated pseudo-samples are used instead of real samples. 
Although several other approaches~\citep{icarl,mer,mir} have been proposed, they still suffer from CF. SPG does not save replay data or generate pseudo-replay data.

\textbf{Parameter isolation-based}:
This approach is most similar to ours SPG.
It tries to learn a sub-network for each task (tasks may share parameters and neurons), which limits knowledge transfer. We have discussed HAT, SupSup, and CAT in \cref{sec:introduction}.
Many others also take similar approaches, e.g., 
Progressive Networks (PGN)~\citep{pgn}, APD~\citep{apd}, PathNet~\citep{pathnet}, PackNet~\citep{packnet}, SpaceNet~\citep{spacenet}, and WSN~\citep{wsn}. 
In particular, PGN allocates a sub-network for each task in advance, and progressively concatenates previous sub-networks while freezing parameters allocated to previous tasks.
APD selectively reuses and dynamically expands the dense network.
Those methods, however, depend on the expansion of network for their performance, which is often not acceptable in cases where many tasks need to be learned.
PathNet splits each layer into multiple sub-modules and finds the best \textit{pathway} designated to each task. 
PackNet freezes important weights for each task by finding them based on pruning.
Although PathNet and PackNet do not expand 
the network along with continual learning, they suffer from  over-consumption of the fixed capacity.
To address this, SpaceNet adopts the sparse training to preserve parameters for future tasks but the performance for each task is sacrificed.
WSN also allocates a subnetwork within a dense network
and selectively reuses subnetworks for previous tasks without expanding the whole network.
Nevertheless, those methods are still limited by the pre-allocated network size as each task monopolizes and consumes some amount of capacity, which results in poorer KT when learning many tasks.

In summary, parameter isolation-based methods suffer from over-consumption of network capacity and have limited KT, which the proposed method tries to address at the same time.

\section{Proposed SPG} 

\begin{figure}[ht]
\centering
\includegraphics[width=.85\hsize]{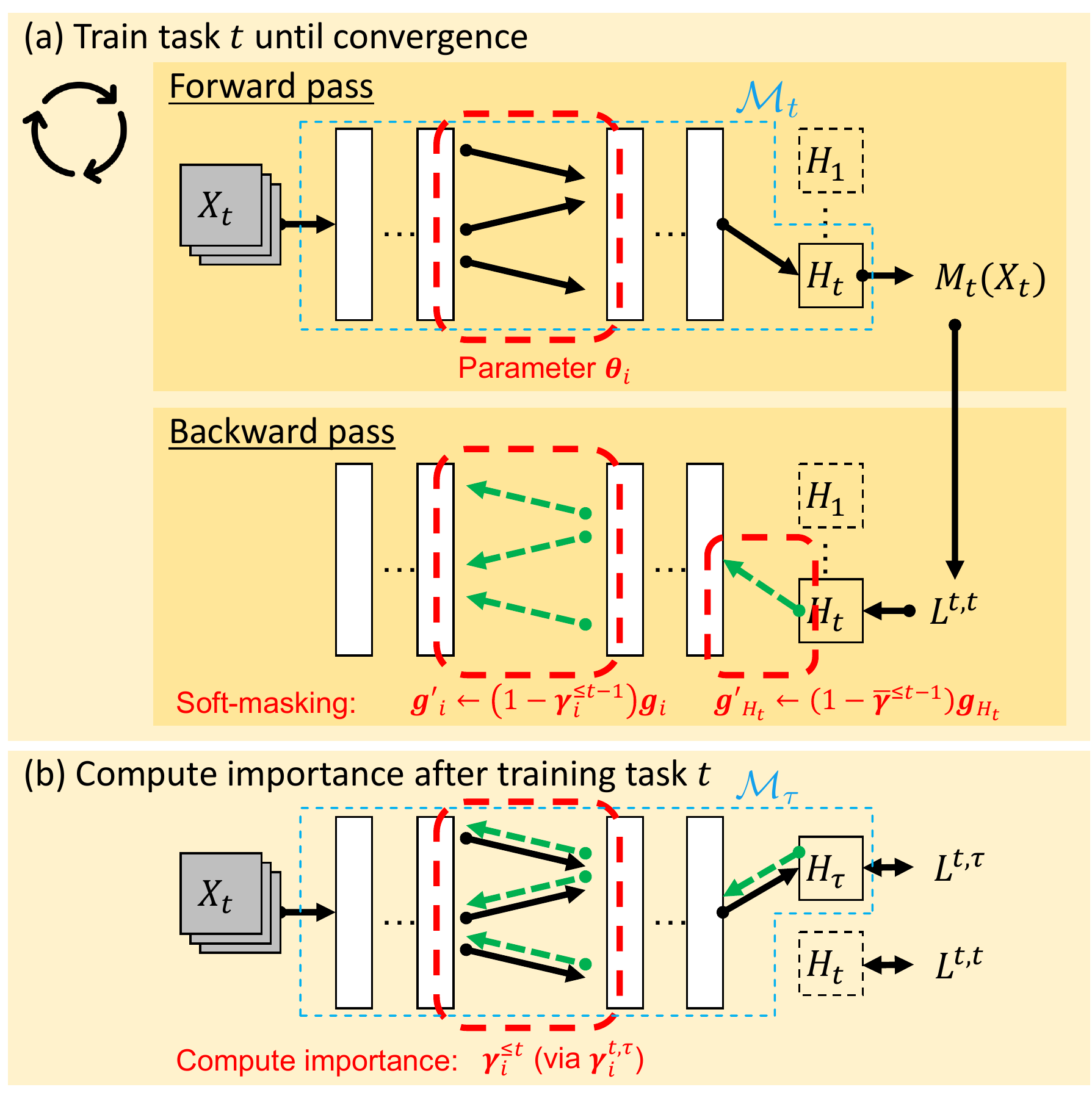}
\caption{
When learning task $t$, SPG proceeds in two steps. Black (solid) and green (dashed) arrows represent forward and backward propagation, respectively. $H_t$ denotes the head for task $t$. 
\textbf{(a)} Training of a model. In the forward pass, nothing extra is done. In the backward pass, the gradients of parameters in the feature extractor $\bm{g}_i$ are changed to $\bm{g}'_i$, 
based on the accumulated importance ($\bm{\gamma}_i^{\le t-1}$). For parameters of the head for task $t$, their gradients $\bm{g}_{H_t}$ are changed to $\bm{g}'_{H_t}$ using the average of accumulated importance ($\bar{\bm{\gamma}}^{\le t-1}$).
\textbf{(b)} Computation of the accumulated importance $\bm{\gamma}_i^{\le t}$. 
}
\label{fig:spg}
\end{figure}

As discussed in \cref{sec:introduction}, the current parameter isolation approaches like HAT~\citep{hat} and SupSup~\citep{supsup} are very effective for overcoming CF, but they hinder knowledge transfer and/or consume too much learning capacity of the network.
For such a model to improve knowledge transfer, it needs to decide which parameters can be shared and updated for a new task. That is the approach taken in CAT~\citep{cat}. CAT  finds similar tasks and removes their masks for updating, but may find wrong similar tasks, which causes CF.
Further, parameters are the atomic information units, not neurons, which HAT masks. If a neuron is masked, all parameters feeding into it are masked, which costs a huge amount of learning capacity. 
SPG directly soft-masks parameters based on their importance to previous tasks, which is a more flexible and uses much less learning space. 
Soft-masking clearly enables automatic knowledge transfer.

\begin{algorithm}[tb]
\caption{
Continual Learning in SPG.
}
\label{alg:spg}
\begin{algorithmic}[1]
\FOR{$t = 1,\cdots,T$}

\STATE \textbf{\# Training of task $t$. $\mathcal{M}_t$ is the model for task $t$ (see \cref{fig:spg}(a)).}

\REPEAT
\STATE Compute gradients $\{ \bm{g}_i \}$ and $\bm{g}_{H_t}$ with $\mathcal{M}_t$ using the task $t$'s data $(\bm{X}_t, \bm{Y}_t)$.

\FORALL{parameters of $i$-th layer}
\STATE $\bm{g}'_i \leftarrow$ \cref{eq:g_i_prime}
\ENDFOR
\FORALL{parameters of the task $t$'s head}
\STATE $\bm{g}'_{H_t} \leftarrow$ \cref{eq:g_ht_prime}
\ENDFOR
\STATE Update $\mathcal{M}_t$ with the modified gradients $\{ \bm{g}'_i \}$ and $\bm{g}'_{H_t}$.

\UNTIL{$\mathcal{M}_t$ converges.}

\STATE \textbf{\# Computing the importance of parameters after training task $t$ (see \cref{fig:spg}(b)).}

\FOR{$\tau = 1, \cdots, t$}
\STATE Compute a loss $\mathcal{L}^{t,\tau}$ in \cref{eq:loss}.
\FORALL{parameters of $i$-th layer}
\STATE $\bm{\gamma}_i^{t,\tau} \leftarrow$ \cref{eq:gamma_i_t_tau}
\ENDFOR
\ENDFOR

\FORALL{parameters of $i$-th layer}
\STATE $\bm{\gamma}_i^t \leftarrow$ \cref{eq:gamma_i_t},\enspace $\bm{\gamma}_i^{\le t} \leftarrow$ \cref{eq:gamma_i_le_t}
\ENDFOR

\STATE Store only $\{ \bm{\gamma}_i^{\le t} \}$ for future tasks.

\ENDFOR
\end{algorithmic}

\end{algorithm}

\cref{fig:spg} and \cref{alg:spg} illustrate how SPG works.
In SPG, the importance of a parameter to a task is computed based on its gradient. We do so because gradients of parameters directly and quantitatively reflect how much changing a parameter affects the final loss.
Additionally, we normalize the gradients of the parameters within each layer to make relative importance more reliable as gradients in different layers can have different magnitude.
The normalized importance scores are accumulated by which the corresponding gradients are reduced in the optimization step to avoid forgetting the knowledge learned from the previous tasks.

\subsection{Computing the Importance of Parameters} \label{sec:computing_gradients}

This procedure corresponds to \cref{fig:spg}(b).
The importance of each parameter to task $t$ is computed right after completing the training of task $t$ following these steps.
Task $t$'s training data, $(\bm{X}_t, \bm{Y}_t)$, is given again to the trained model of task $t$, and the gradient of each parameter in $i$-th layer (i.e., each weight or bias of each layer) is then computed and used for computing the importance of the parameter. Note that we use $\bm{\theta}_i$ (a vector) to represent all parameters of the $i$-th layer. This process does not update the model parameters. 
The reason that the importance is computed after training of the current task has converged is as follows. Even after a model converges, some parameters can have larger gradients, which indicate that changing those parameters may take the model out of the (local) minimum leading to forgetting. On the contrary, if all parameters have similar gradients (i.e, balanced directions of gradients), changing the parameters will not likely to change the model much to cause forgetting.
Based on this assumption, we utilize the normalized gradients after training as a signal to indicate such dangerous parameter updates.
The proposed mechanism in SPG has the merit that it keeps the model flexible as it does not fully block parameters using an importance threshold or binary masks.
While HAT completely blocks important neurons, which results in the loss of trainable parameters over time, SPG allows most parameters remain ``alive'' even when most of them do not change much.

\begin{figure}[tb]
\centering
\includegraphics[width=.77\hsize]{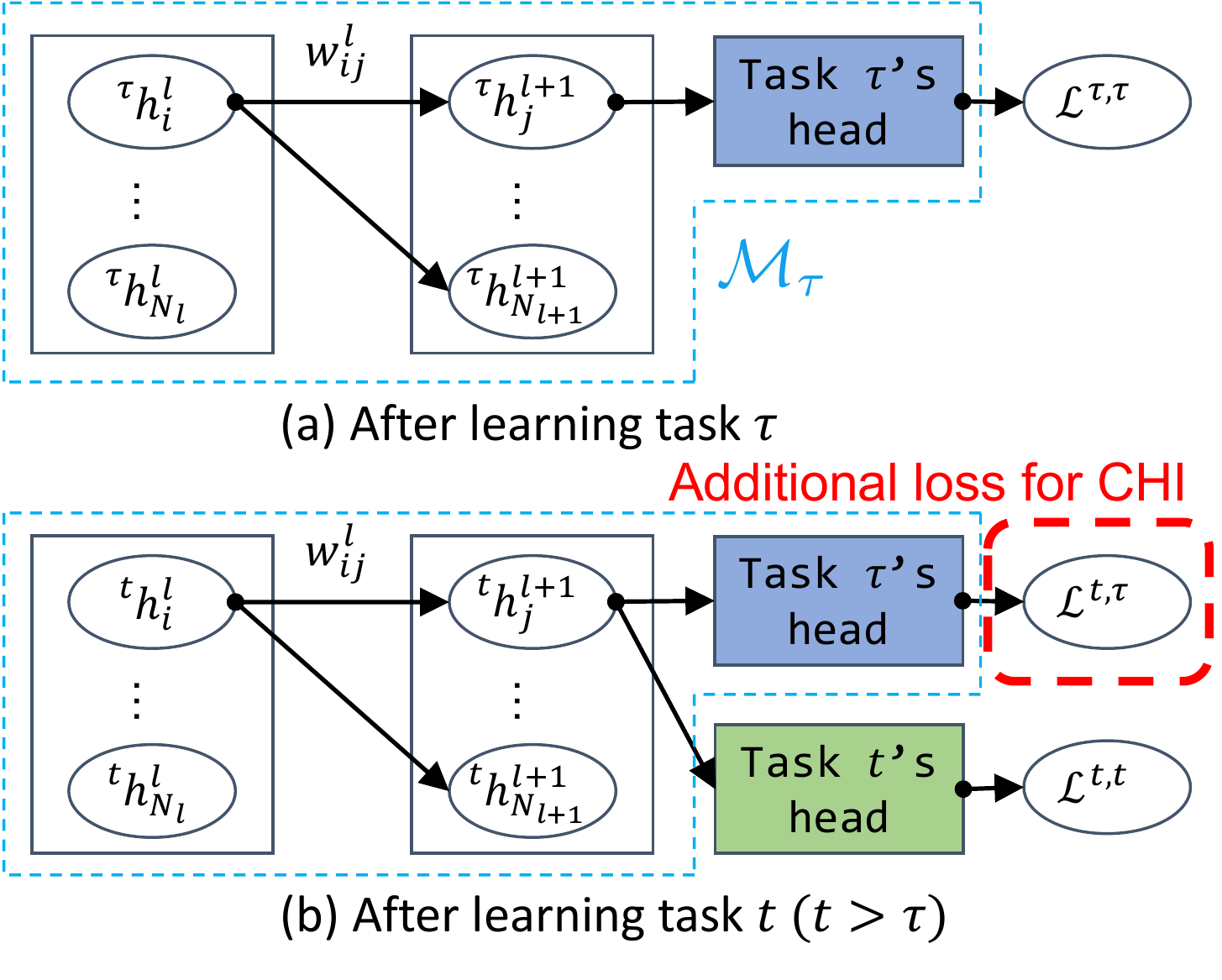}
\caption{
Cross-head importance (CHI).
In the above figures, ${}^th_i^l$ and $w_{ij}^l$ represents the output of the $i$-th neuron in the $l$-th layer just after training task $t$ and the parameter in $l$-th layer connecting between the neurons ${}^th_i^l$ to ${}^th_j^{l+1}$, respectively.
\textbf{(a)} The importance of $w_{ij}^l$ to task $\tau$ is computed based on its gradient, ${\partial \mathcal{L}^{\tau,\tau}}/{\partial w_{ij}^{l}}$ and then accumulated.
\textbf{(b)} After learning task $t$ ($t > \tau$), the state of related parameters might have been changed. To reflect importance to task $\tau$ again with the current neurons' output (e.g., ${}^th_i^l$ rather than old ${}^\tau h_i^l$), an additional loss, $\mathcal{L}^{t,\tau}$, is computed at task $\tau$'s head using task $t$'s data as unlabeled data for task $\tau$.
}
\label{fig:chi}
\vspace{-5mm}
\end{figure}

Additionally, computing the gradients based only on the model ($\mathcal{M}_t$) of the current task $t$ does not deal with another issue. 
We use an example illustrated in \cref{fig:chi}.
For example, just after learning task $\tau$, the gradient of a parameter is computed and normalized among the same layer's parameters to be accumulated.
Even though during learning task $t$ ($t > \tau$), the parameter is not much changed considering its accumulated importance, at the end of learning task $t$, the state of related parameters might have been changed, by which the normalized importance may become less useful. To reflect the parameter's importance to task $\tau$ again in the current network state, we introduce \textit{cross-head importance} (CHI) mechanism. In particular, an additional loss, $\mathrm{Sum}(\mathcal{M}_\tau(\bm{X}_t))$, is computed with each previous task's head by substituting task $t$'s data as unlabeled data for the previous tasks. By taking this loss, parameters affecting the logits more  for previous tasks are regarded more important.
Finally, both the normalized importance computed for the current task's head and the ones for previous tasks' heads in CHI are considered by taking element-wise maximum, as shown in \cref{eq:gamma_i_t}.

To put things together, the proposed method computes the normalized importance, $\bm{\gamma}_i^t$, of the parameters of the $i$-th layer, $\bm{\theta}_i$, using each task $\tau$'s model ($1\!\le\!\tau\!\le\!t$), $\mathcal{M}_\tau$. 
\begin{gather}
\bm{\gamma}_i^{t, \tau} = \left| \tanh{\left( \mathrm{Norm} \left( \frac{\partial \mathcal{L}^{t, \tau}}{\partial \bm{\theta}_i} \right) \right)} \right|, \label{eq:gamma_i_t_tau} \\
\mathcal{L}^{t, \tau} = 
\begin{cases}
\mathcal{L}\left( \mathcal{M}_\tau\left(\bm{X}_t\right), \bm{Y}_t \right) & (\tau = t) \\
\mathrm{Sum}\left( \mathcal{M}_\tau\left(\bm{X}_t\right) \right) & (\tau < t)
\end{cases}, \label{eq:loss} \\
\mathrm{Norm}(\bm{x}) =  \frac{\bm{x} - \mathrm{Mean}(\bm{x})}{\sqrt{\mathrm{Var}(\bm{x})}}, \label{eq:norm} \\
\bm{\gamma}_i^t = \max{\left( \bm{\gamma}_i^{t,1}, \cdots, \bm{\gamma}_i^{t,t} \right)}, \label{eq:gamma_i_t}
\end{gather}
where $\max{(\cdot)}$ and $\mathcal{L}$ mean element-wise maximum and a loss function, respectively. 
\cref{eq:gamma_i_t_tau} normalizes the gradients over the same layer to avoid the discrepancies caused by large differences of gradient magnitudes in different layers.
For the current task's head (i.e., $\tau\!=\!t$), a normal loss function (e.g., cross entropy) is used as $\mathcal{L}^{t,t}$ in \cref{eq:loss}. 
However, for each previous task’s head (when $\tau\!<\!t$), since the current task data do not belong to any previous classes, 
the loss $\mathcal{L}^{t,\tau}$ is defined by $\mathrm{Sum}(\mathcal{M}_\tau(\bm{X}_t))$ over previous classes' logits in the proposed CHI mechanism. 
Essentially, this operation computes the importance of parameters based on the data of task $t$'s impact on all tasks learned so far.
Finally, to prevent forgetting as much as possible, we take the accumulated importance, $\bm{\gamma}_i^{\le t}$, as follows:
\begin{gather}
\bm{\gamma}_i^{\le t} = \max{\left( \bm{\gamma}_i^t, \bm{\gamma}_i^{\le t-1} \right)}, \label{eq:gamma_i_le_t}
\end{gather}
where an all-zero vector is used as $\bm{\gamma}_i^{\le 0}$. This $\bm{\gamma}_i^{\le t}$ depicts how important each parameter is to all the learned tasks.

\textbf{Memory needed to save parameter importance}: 
Regardless of the number of tasks, at any time only the accumulated importance $\bm{\gamma}_i^{\le t}$ is saved after the learning of each task so that it can be used again in the next task for \cref{eq:gamma_i_le_t}. $\bm{\gamma}_i^{\le t}$ has the same size as the number of parameters, $|\bm{\theta}_i|$.

\subsection{Soft-Masking of Feature Extractor} \label{sec:training_phase}
This procedure appears in \cref{fig:spg}(a).
To suppress the update of important parameters in the backward pass in learning task $t$, the gradients of all parameters in the shared feature extractor are modified (i.e., \textit{\textbf{soft-masked}}) based on the accumulated importance as follows (i.e., each parameter is soft-masked by a different amount according to its accumulated importance): 
\begin{gather}
\bm{g}_i' = \left( 1 - \bm{\gamma}_i^{\le t-1} \right) \bm{g}_i, \label{eq:g_i_prime}
\end{gather}
where $\bm{g}_i$ and $\bm{g}_i'$ represent the original gradients of the parameters of the $i$-th layer (i.e., $\bm{\theta}_i$) and the modified ones, which will be used in the actual optimization, respectively.

\subsection{Soft-Masking of Classification Head}
We found that the above soft-masking may induce another problem.
If only the feature extractor's parameters are soft-masked, the model will try to find an optimal solution mainly by updating the classification head since its parameters are not masked and thus can be updated more easily than the feature extractor. However, this discourages the learning of the feature extractor, which hinders knowledge transfer.

To achieve a balanced training of the feature extractor and the classification head,
we need to \textbf{slow down} the learning of the head by reducing the gradients of the head's parameters based on how much the feature extractor's parameters are soft-masked. 
We still use the soft-masking idea, but all parameters in the head are soft-masked by an equal amount.
Specifically, the gradients of all parameters in task $t$'s head, $\bm{g}_{H_t}$, are soft-masked by the average ($\bar{\bm{\gamma}}^{\le t-1}$) of the accumulated importance of all the parameters in the feature extractor (i.e., $\{ \bm{\gamma}_i^{\le t-1} \}$). The modified gradients, $\bm{g}_{H_t}'$, are used in optimization. 
 \begin{gather}
\bm{g}_{H_t}' = \left( 1 - \bar{\bm{\gamma}}^{\le t-1} \right) \bm{g}_{H_t} \label{eq:g_ht_prime}
 \end{gather}
 
Note that SPG has no specific hyper-parameter and does not employ anything special in the forward pass except it needs the task id to locate the correct head, which follows the standard TIL scenario.  
To our knowledge, neither soft-masking of the parameters in the feature extractor nor in the classification head has been done by any existing work.

\section{Experiments} \label{sec:experiments}

\textbf{Datasets}: 
The proposed SPG is evaluated using 5 CL datasets. Their statistics are given in \cref{tab:statisctics_datasets}. Below, we use ``-$n$'' to depict that $n$ tasks are created from each dataset ($n$ takes 10 or 20). 
Classes in the first three datasets are split by task, so each task has a disjoint set of classes. On the other hand, all tasks in the last two datasets have the same set of classes.
We refer to the tasks in the former datasets as ``dissimilar tasks'' in which CF is the essential problem to solve, while we regard the tasks in latter datasets as ``similar tasks'' as the ability of knowledge transfer is more important.

\textbf{(1) CIFAR100-$n$ (C-$n$)}:
CIFAR100~\citep{cifar100} is a dataset that has images of 100 classes. We split it into $n$ tasks so that each task has $100/n$ classes. 
\textbf{(2) TinyImageNet-$n$ (T-$n$)}:
TinyImageNet~\citep{tinyimagenet} is a modified subset of the original ImageNet~\citep{imagenet} dataset, and has 200 classes. Each task contains $200/n$ classes.
\textbf{(3) ImageNet-$100$ (I-$100$)}:
ImageNet~\citep{imagenet} contains 1000 classes of objects. We split it to 100 tasks, each of which has 10 classes, to stress-test systems using a large number of tasks and classes.
\textbf{(4) F-CelebA-$n$ (FC-$n$)}:
Federated CelebA~\citep{celebafemnist} is a dataset of celebrities' face images with several attributes. We use it with binary labels indicating whether he/she in the image is smiling or not. Each task consists of images of one celebrity. 
\textbf{(5) F-EMNIST-$n$ (FE-$n$)}:
Federated EMNIST~\citep{celebafemnist} is a dataset that has 62 classes of hand-written symbols written by different persons. Each task consists of hand-written symbols of one person.

\begin{table}[t]

\vspace{-3mm}

\caption{Statistics of the CL datasets. $n$ can take $10$ and $20$.
Validation sets are used for early stopping. 
}
\label{tab:statisctics_datasets}
\centering
\resizebox{.95\hsize}{!}{
\begin{tabular}{crrrrr}
\toprule
Dataset & \multicolumn{1}{c}{\#Tasks} & \multicolumn{1}{c}{\#Classes per task} & \multicolumn{1}{c}{\#Train} & \multicolumn{1}{c}{\#Validation} & \multicolumn{1}{c}{\#Test} \\ \midrule
C-$n$ & $n$ & $100/n$ & $45,000$ & $5,000$ & $10,000$ \\
T-$n$ & $n$ & $200/n$ & $90,000$ & $10,000$ & $10,000$ \\
I-100 & $100$ & $10$ & $1,000,000$ & $100,000$ & $50,000$ \\
FC-$n$ & $n$ & $2$ & $400n$ & $40n$ & $80n$ \\
FE-$n$ & $n$ & $62$ & $3100n$ & $310n$ & $620n$ \\
\bottomrule
\end{tabular}
}
\vspace{-5mm}
\end{table}

\textbf{Baselines}:
We use 16 baselines. 11 of them are existing classical and most recent task incremental learning (TIL) systems, \textbf{EWC}~\citep{ewc}, \textbf{A-GEM}~\citep{a-gem},  \textbf{SI}~\citep{si}, \textbf{UCL}~\citep{ucl}, \textbf{TAG}~\citep{tag}, \textbf{PGN}~\citep{pgn}, \textbf{PathNet}~\citep{pathnet}, \textbf{HAT}~\citep{hat}, \textbf{CAT}~\citep{cat}, \textbf{SupSup}~\citep{supsup}, and \textbf{WSN}~\citep{wsn}.
Additionally, three simple methods are used for references: multi-task learning (\textbf{MTL}) that trains all the tasks together,
one task learning (\textbf{ONE}) that learns a separate model/network for each task and thus has no CF or KT,
and naive continual learning (\textbf{NCL})  
that learns each new task without taking any care of previous tasks, i.e., no mechanism to deal with CF. 
Since HAT, our main baseline and perhaps the most effective TIL system, adopts AlexNet~\citep{alexnet} as its backbone, all our experiments are conducted with AlexNet. For other baselines, their original codes are used with switching their backbones to AlexNet for fair comparison. 
Furthermore,
to compare our \textbf{\textit{soft-masking}} with the \textbf{\textit{regularization-based approach}} and our gradient-based importance with Fisher information matrix (FI) based importance in EWC, two more baselines \textbf{EWC-GI} and \textbf{SPG-FI} are created. EWC-GI is EWC with its FI based importance replaced by our gradient-based importance (GI) in \cref{sec:computing_gradients}, i.e., the same penalty/regularization in EWC is applied on our accumulated importance, $\bm{\gamma}_i^{\le t-1}$ in \cref{eq:gamma_i_le_t} when learning task $t$ (no soft-masking). SPG-FI is SPG with our gradient-based importance replaced by FI based importance in EWC. 
The network size of each method is presented in \cref{sec:app_network_size}.

\begin{table*}[t]
\centering
\caption{
Accuracy results in percent.
Best methods in each dataset are emphasized in \textbf{bold}, and second best methods are \underline{underlined}. 
}
\label{tab:results_acc}
\resizebox{.95\hsize}{!}{
\begin{tabular}{crrrrrcrrrrc}
\toprule
& \multicolumn{6}{c}{Dissimilar tasks} & \multicolumn{5}{c}{Similar tasks} \\ \cmidrule(lr){2-7} \cmidrule(lr){8-12}
Model &
\multicolumn{1}{c}{C-10} & \multicolumn{1}{c}{C-20} & \multicolumn{1}{c}{T-10} & \multicolumn{1}{c}{T-20} & \multicolumn{1}{c}{I-100} & \textbf{(Avg.)} &
\multicolumn{1}{c}{FC-10} & \multicolumn{1}{c}{FC-20} & \multicolumn{1}{c}{FE-10} & \multicolumn{1}{c}{FE-20} & \textbf{(Avg.)} \\

\midrule

(MTL) &
$76.4\std{0.3}$ & $78.4\std{0.4}$ & 
$52.7\std{0.3}$ & $59.6\std{1.2}$ &
$64.8\std{0.4}$ &
$66.4$ &
$87.5\std{0.7}$ & $88.3\std{0.2}$ &
$86.2\std{0.7}$ & $87.2\std{2.2}$ &
$87.3$ \\

(ONE) &
$66.9\std{3.1}$ & $76.5\std{0.7}$ &
$43.5\std{3.0}$ & $54.5\std{0.9}$ &
$49.3\std{0.4}$ &
$58.1$ &
$74.6\std{2.6}$ & $78.8\std{2.3}$ &
$80.9\std{1.5}$ & $79.7\std{1.4}$ &
$78.5$ \\

\hdashline

NCL &
$50.9\std{1.7}$ & $54.2\std{5.3}$ &
$37.2\std{0.9}$ & $41.1\std{1.0}$ & 
$30.6\std{1.2}$ &
$42.8$ &
$84.4\std{1.8}$ & $84.1\std{1.4}$ &
$86.1\std{0.8}$ & $86.4\std{0.3}$ &
$85.3$ \\

A-GEM &
$50.8\std{1.0}$ & $56.9\std{7.1}$ & 
$36.2\std{0.6}$ & $41.7\std{1.1}$ & 
$32.1\std{1.1}$ &
$43.5$ &
$83.2\std{3.6}$ & $83.2\std{1.9}$ &
$86.6\std{0.2}$ & $86.9\std{0.3}$ &
$85.0$ \\

PGN &
$65.1\std{0.6}$ & $75.5\std{0.4}$ &
$44.0\std{0.8}$ & $53.5\std{0.4}$ &
$45.2\std{0.4}$ &
$56.7$ &
$74.7\std{3.5}$ & $74.7\std{2.8}$ &
$82.5\std{1.0}$ & $82.6\std{0.3}$ &
$78.6$ \\

PathNet &
$69.1\std{0.5}$ & $75.5\std{1.0}$ &
$46.0\std{1.5}$ & $51.9\std{1.6}$ &
$42.0\std{1.5}$ &
$56.9$ &
$79.3\std{1.2}$ & $80.5\std{0.4}$ &
$84.3\std{0.2}$ & $84.5\std{0.4}$ &
$82.1$ \\

HAT &
$62.8\std{0.7}$ & $71.8\std{1.1}$ &
$45.5\std{1.0}$ & $51.7\std{2.1}$ &
$45.3\std{1.9}$ &
$55.4$ &
$79.0\std{3.1}$ & $81.9\std{0.7}$ &
$83.8\std{0.9}$ & $84.6\std{0.8}$ &
$82.3$ \\

CAT &
$64.2\std{0.6}$ & $73.9\std{1.1}$ &
$43.7\std{0.6}$ & $50.9\std{0.8}$ & 
\multicolumn{1}{c}{N/A} &
N/A &
$82.9\std{1.3}$ & $82.9\std{3.7}$ &
$82.9\std{1.4}$ & $84.1\std{0.9}$ &
$83.2$ \\

SupSup &
$66.2\std{0.2}$ & $75.6\std{0.3}$ &
$44.0\std{0.2}$ & $54.1\std{0.3}$ &
$48.6\std{0.1}$ &
$57.7$ &
$75.0\std{2.3}$ & $78.1\std{1.4}$ &
$80.5\std{0.5}$ & $79.7\std{0.2}$ &
$78.3$ \\

UCL &
$64.8\std{0.8}$ & $74.0\std{0.6}$ &
$45.4\std{0.3}$ & $55.1\std{0.5}$ &
$37.4\std{0.6}$ &
$55.4$ &
$86.2\std{0.5}$ & $86.5\std{0.5}$ &
$85.1\std{0.7}$ & $85.0\std{1.6}$ &
$85.7$ \\

SI &
$62.9\std{0.3}$ & $70.3\std{0.7}$ &
$45.9\std{0.6}$ & $52.6\std{0.7}$ &
$44.1\std{0.2}$ &
$55.2$ &
$86.4\std{0.8}$ & $\underline{86.8}\std{0.3}$ &
$\underline{87.6}\std{0.3}$ &  $\textbf{87.9}\std{0.2}$ &
$\underline{87.2}$ \\

TAG &
$60.6\std{0.7}$ & $68.4\std{0.9}$ &
$43.0\std{0.8}$ & $49.5\std{0.4}$ &
$44.9\std{0.3}$ &
$53.3$ &
$74.3\std{3.7}$ & $77.3\std{2.1}$ &
$84.2\std{0.5}$ & $83.8\std{0.4}$ &
$79.9$ \\

WSN &
$\textbf{69.3}\std{0.2}$ & $\textbf{76.9}\std{0.5}$ &
$47.8\std{0.5}$ & $\underline{57.8}\std{0.5}$ &
$51.9\std{0.4}$ &
$\underline{60.7}$ &
$83.9\std{1.2}$ & $84.1\std{0.7}$ &
$85.5\std{0.2}$ & $86.3\std{0.2}$ &
$84.9$ \\

EWC &
$61.6\std{0.9}$ & $60.7\std{2.7}$ & 
$36.5\std{1.1}$ & $41.5\std{1.3}$ & 
$25.4\std{1.4}$ &
$45.1$ &
$81.2\std{3.0}$ & $86.1\std{0.9}$ &
$86.9\std{0.3}$ & $86.8\std{0.6}$ &
$85.2$ \\

\hdashline

EWC-GI &
$63.3\std{1.2}$ & $60.1\std{1.9}$ &
$\underline{48.3}\std{1.0}$ & $48.6\std{1.9}$ &
$\underline{52.7}\std{0.2}$ &
$54.6$ &
$83.4\std{3.0}$ & $84.6\std{1.7}$ &
$86.6\std{1.5}$ & $\underline{87.4}\std{0.9}$ &
$85.5$ \\

SPG-FI &
$60.5\std{0.2}$ & $67.7\std{1.0}$ & 
$43.9\std{0.6}$ & $51.2\std{0.8}$ & 
$48.8\std{0.8}$ & 
$54.4$ &
$\underline{86.7}\std{0.6}$ & $86.2\std{0.8}$ & 
$87.5\std{0.3}$ & $\textbf{87.9}\std{0.2}$ & 
$87.1$ \\

\midrule

\textbf{SPG} &
$\underline{67.7}\std{0.3}$ & $\underline{75.9}\std{1.1}$ & 
$\textbf{48.4}\std{0.3}$ & $\textbf{59.1}\std{0.5}$ & 
$\textbf{58.1}\std{0.5}$ & 
$\textbf{61.8}$ &
$\textbf{87.0}\std{0.9}$ & $\textbf{87.1}\std{0.2}$ & 
$\textbf{87.7}\std{0.2}$ & $\textbf{87.9}\std{0.1}$ & 
$\textbf{87.4}$ \\

\bottomrule
\end{tabular}
}
\vspace{-3mm}
\end{table*}

\textbf{Evaluation Metrics}:
The following three metrics are used. Let $\alpha_i^j$ be the test accuracy of task $i$ task just after a model completes task $j$. 

\textbf{(1)} \textit{Accuracy}:
The average of accuracy for all tasks of a dataset after learning the final task. It is computed by $1/T \sum_t^T \alpha_t^T$, where $T$ is the total number of tasks. \\ 
\textbf{(2)} \textit{Forward transfer}:
This measures how much the learning of previous tasks contributes to the learning of the current task. It is computed by $1/T \sum_t^T \left( \alpha_t^t - \beta_t \right)$, where $\beta_t$ represents the test accuracy of task $t$ in the ONE method, which learns each task separately. \\
\textbf{(3)} \textit{Backward transfer}:
This measures how the learning of the current task affects the performance of the previous tasks. Negative values indicate forgetting. It is computed by $1/(T-1) \sum_t^{T-1} \left( \alpha_t^T - \alpha_t^t \right)$.

\subsection{Training Details} \label{sec:training_details}
The networks are trained with SGD by minimizing the cross-entropy loss except for TAG, which uses the RMSProp optimizer as SGD-based TAG has not been provided by the authors.
The mini-batch size is 64 except MTL that uses 640 for its stability to learn more tasks and classes together.
Hyper-parameters, such as the dropout rate or each method's specific hyper-paramters, are searched based on Tree-structured Parzen Estimator~\citep{tpe}.
With the found best hyper-parameters, the experiments are conducted 5 times with different seeds, and the average accuracy and standard deviation are reported.

\subsection{Results} \label{sec:results}

\cref{tab:results_acc,tab:results_fwt,tab:results_bwt} report the accuracy, forward and backward transfer results, respectively. 
Since CAT takes too much time to train, proportionally to the square of the number of tasks, we are unable to get its results for I-100 (ImageNet with 100 tasks) due to our limited computational resources.

\textbf{Dissimilar Tasks} (C-$n$, T-$n$, I-$100$):
MTL performs the best in all cases, but NCL performs poorly due to serious CF (negative backward transfer) as expected. While PGN, PathNet, HAT, CAT, and SupSup can achieve training with no forgetting (0 backward transfer), on average SPG clearly outperforms all of them as their transfer is limited. Although SPG  underperforms PathNet in C-10, its accuracy is markedly lower in the other settings due to PathNet's capacity problem (see \cref{sec:capacity}).
The backward transfer results in \cref{tab:results_bwt} show that SPG has slight forgetting. However, its positive forward transfer results in \cref{tab:results_fwt} more than make up for the forgetting and give SPG the best final accuracy in most cases. 
As we explained in \cref{sec:related_work}, the regularization-based approach is closely related to our work. However, the representative methods, EWC, SI and UCL, perform poorly due to serious CF (see \cref{tab:results_bwt}) that cannot be compensated by their positive forward transfer (see \cref{tab:results_fwt}). 
Although TAG has almost no CF, its forward transfer is limited, resulting in poorer final accuracy.
While SPG underperforms WSN in C-10 and C-20 slightly, SPG outperforms all baselines in other dissimilar task datasets; especially, in the more realistic and difficult dataset I-100, for which SPG is significantly better, outperforming WSN by 6.2\%.

Comparing our gradient based importance (GI) and Fisher information (FI) matrix based importance, we observe that EWC-GI outperforms EWC except for C-20 (EWC is less than 1\% better), and SPG significantly outperforms SPG-FI in all cases. 
Comparing soft-masking and regularization using the same importance measure, we can see SPG is markedly better than EWC-GI, and SPG-FI is also better than EWC except for C-10 (EWC is only 1\% better).
\textcolor{black}{
These results clearly demonstrate that our gradient based importance (GI) and soft-masking are much more effective than standard \textit{\textbf{regularization methods}}.
}

\textbf{Similar Tasks} (FC-$n$, FE-$n$):
\cref{tab:results_acc} shows that SPG achieves the best in all cases due to its strongest \textit{\textbf{positive forward}} (see \cref{tab:results_fwt}) and \textit{\textbf{positive backward knowledge transfer}} ability (see \cref{tab:results_bwt}).
NCL, A-GEM, UCL, and SI also perform well with positive or very little negative backward transfer since the tasks are similar and CF hardly happens.
TAG underperforms them due to it's limited transfer.
PGN, PathNet, HAT, and WSN
have lower accuracy as their forward transfer is limited (i.e., they just reuse learned parameters in the forward pass) and no positive backward transfer.
SupSup, which does not have any mechanism for knowledge transfer, results in much lower performance.
CAT is slightly better due to its stronger positive forward transfer and no negative backward transfer.

Comparing GI and FI, SPG-FI and EWC-GI perform similarly to SPG and EWC as suppressing updates of important parameters becomes less critical for similar tasks and thus the choice of GI or FI is not important.
Comparing soft-masking and regularization, EWC-GI and EWC are worse than SPG and SPG-FI in all cases, indicating that soft-masking is still more effective as regularization may hinder the learning of new tasks more.

\textbf{Summary}:
SPG markedly outperforms all the baselines. 
When tasks are dissimilar, its positive forward transfer capability overcomes its slight forgetting (negative backward transfer) and achieves the best overall results. It has the strong positive forward transfer even with dissimilar tasks, which has not been realized by the other parameter isolation-based baselines.
When tasks are similar, SPG has both positive forward and backward transfer to achieve the best accuracy results. Keeping most parameters trainable in SPG promotes knowledge transfer. Moreover, we observe that soft-masking (SPG and SPG-FI) is better than regularization (EWC-GI and EWC) and that our gradient-based importance (SPG and EWC-GI) is better than FI (SPG-FI and EWC).

\subsubsection{Capacity Consumption} \label{sec:capacity}
The reason that PGN, PathNet, HAT, CAT, and WSN
have lower accuracy on average than SPG despite the fact that they can learn with no forgetting (see backward transfer in \cref{tab:results_bwt}) is mainly because they suffer from the capacity problem, which is also indirectly reflected in lower forward transfer in \cref{tab:results_fwt}.
Although SupSup does not suffer from this problem, its architecture prevents transfer and gives markedly poorer performances especially in similar tasks (see \cref{tab:results_acc}).
As discussed in \cref{sec:introduction}, since these parameter isolation-based methods freeze a sub-network for each task, as more tasks are learned, the capacity of the network left for learning new knowledge becomes less and less 
except for SupSup,
which leads to poorer performance for later tasks. 
\cref{tab:blocking_ratio} shows the percentage of parameters in the whole network that are completely blocked by HAT and SPG (layer-wise results are given in \cref{sec:app_capacity_layerwise}). 
SPG blocks much fewer parameters than HAT does in all cases. This advantage allows SPG to have more flexibility and capacity to learn while mitigating forgetting, which leads to better performances.

\begin{figure}[t]

\begin{minipage}{.48\hsize}
\centering
\includegraphics[width=\hsize]{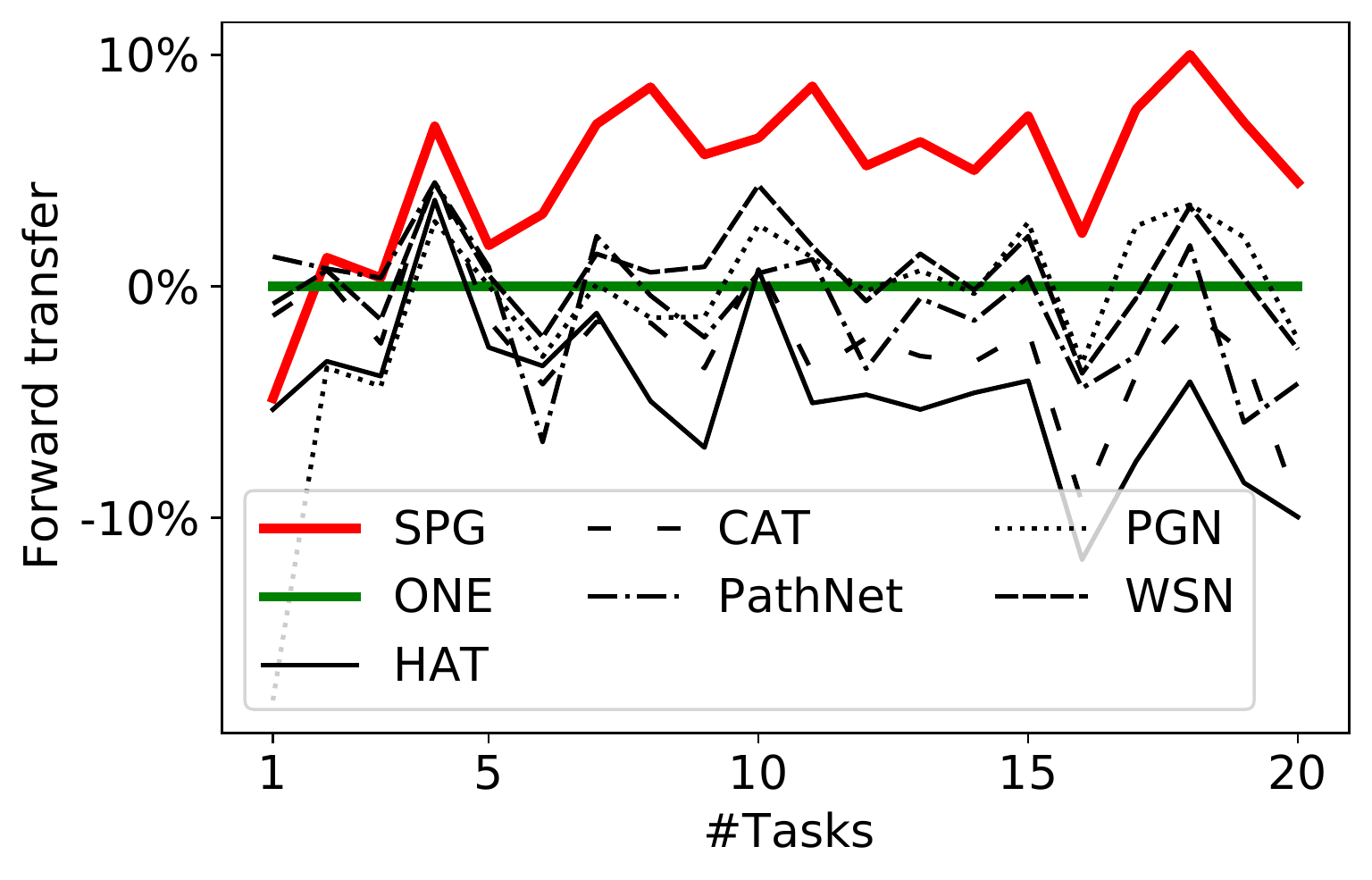}
\subcaption{C-20}
\end{minipage}
\hfill
\begin{minipage}{.48\hsize}
\centering
\includegraphics[width=\hsize]{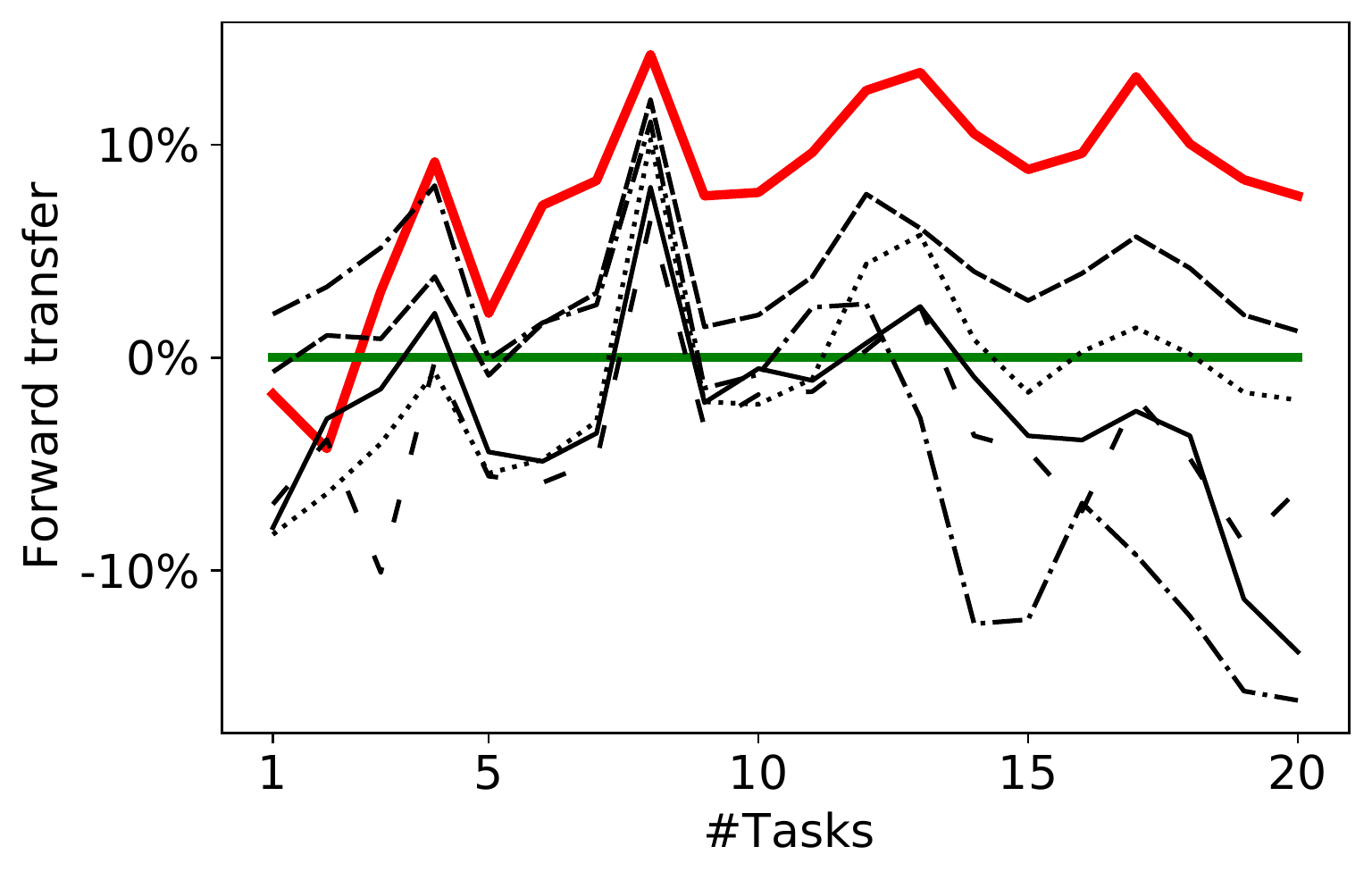}
\subcaption{T-20}
\end{minipage}

\medskip

\begin{minipage}{.48\hsize}
\centering
\includegraphics[width=\hsize]{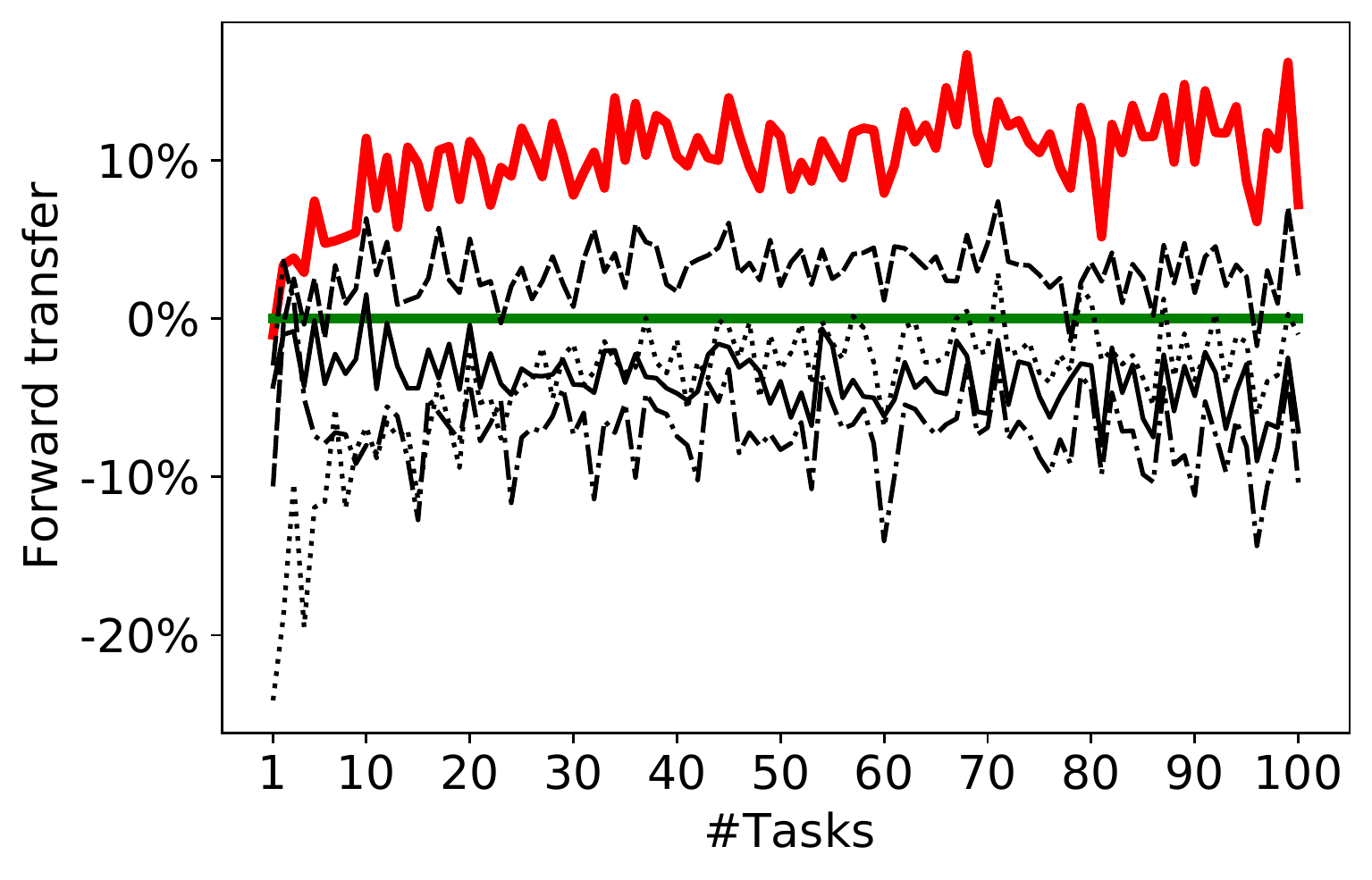}
\subcaption{I-100}
\end{minipage}
\hfill
\begin{minipage}{.48\hsize}
\centering
\includegraphics[width=\hsize]{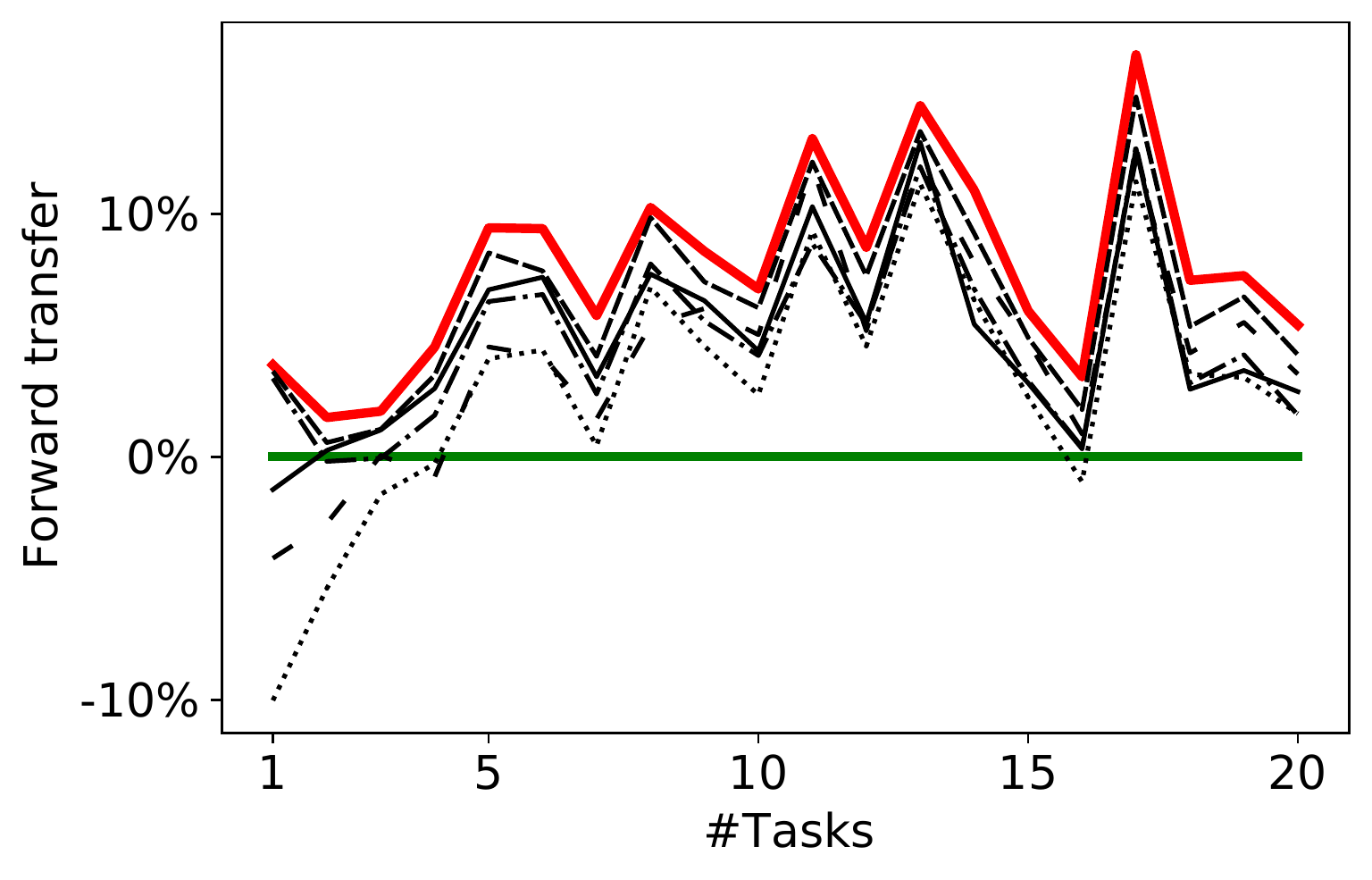}
\subcaption{FE-20}
\end{minipage}

\caption{
The forward transfer of each task along with the number of tasks learned. 
(a) to (c) are the plots with dissimilar tasks, while (d) is the one with similar tasks.
}
\label{fig:results_fwt}

\vspace{-3mm}

\end{figure}

\cref{fig:results_fwt} plots the forward transfer of the limited datasets due to the space limitations (all results are presented in \cref{sec:app_results_fwt}). 
A positive value in the figures means that a method's forward test result (the test result of the task obtained right after the task is learned) is better than ONE, benefited by the forward knowledge transfer from previous tasks.
We can clearly observe a downward trend of these baselines for dissimilar tasks ((a)-(c)).
We believe that this is due to the capacity problem, i.e., as more tasks are learned, they gradually lose their learning capacity. 
On the other hand, SPG shows a upward trend in all cases, even for dissimilar tasks, thanks to its positive forward transfer, which indicates SPG has a much higher capacity to learn.
For \cref{fig:results_fwt}(d), the difference is not obvious as the tasks are similar and the capacity problem is less serious.
However, SPG still keeps the best forward transfer.

\begin{table}[t]
\centering
\caption{
Forward transfer results in percent. 
}
\label{tab:results_fwt}
\resizebox{\hsize}{!}{
\begin{tabular}{crrrrrrrrrrr}
\toprule
& \multicolumn{6}{c}{Dissimilar tasks} & \multicolumn{5}{c}{Similar tasks} \\ \cmidrule(lr){2-7} \cmidrule(lr){8-12}
Model &
\multicolumn{1}{c}{C-10} & \multicolumn{1}{c}{C-20} & \multicolumn{1}{c}{T-10} & \multicolumn{1}{c}{T-20} & \multicolumn{1}{c}{I-100} & \multicolumn{1}{c}{(Avg.)} &
\multicolumn{1}{c}{FC-10} & \multicolumn{1}{c}{FC-20} & \multicolumn{1}{c}{FE-10} & \multicolumn{1}{c}{FE-20} & \multicolumn{1}{c}{(Avg.)} \\
\midrule

NCL &
$-7.1$ & $-2.7$ &
$-3.2$ & $-4.3$ & 
$-0.7$ &
$-3.6$ &
$+7.9$ & $\underline{+6.2}$ &
$+4.1$ & $+5.9$ &
$+6.0$ \\

A-GEM &
$-3.2$ & $-0.9$ & 
$-3.7$ & $-3.5$ & 
$-0.5$ &
$-2.4$ &
$+8.4$ & $+5.4$ &
$+4.3$ & $+6.4$ &
$+6.1$ \\

PGN &
$-1.8$ & $-1.0$ & 
$+0.5$ & $-1.0$ & 
$-3.9$ &
$-1.4$ &
$+0.2$ & $-4.1$ & 
$+1.6$ & $+2.9$ &
$+0.1$ \\

PathNet &
$+2.2$ & $-0.9$ & 
$+2.5$ & $-2.6$ & 
$-7.1$ &
$-1.2$ &
$+4.7$ & $+1.7$ & 
$+3.4$ & $+4.8$ &
$+3.7$ \\

HAT & 
$-4.0$ & $-4.6$ & 
$+2.0$ & $-2.8$ &
$-3.8$ &
$-2.7$ &
$+4.4$ & $+3.2$ &
$+2.9$ & $+4.9$ & 
$+3.8$ \\

CAT &
$-2.7$ & $-2.5$ &
$+0.1$ & $-3.6$ &
N/A &
N/A &
$+8.4$ & $+4.1$ &
$+2.0$ & $+4.4$ & 
$+4.7$ \\

SupSup &
$-0.6$ & $-0.8$ &
$+0.5$ & $-0.3$ &
$-0.4$ &
$-0.4$ &
$+0.5$ & $-0.8$ &
$-0.4$ & $+0.1$ &
$-0.2$ \\

UCL &
$+3.9$ & $\textbf{+6.1}$ &
$+4.9$ & $+7.8$ &
$+2.7$ &
$+5.1$ &
$+8.3$ & $+5.6$ &
$+3.0$ & $+4.7$ &
$+5.4$ \\

SI &
$\textbf{+7.5}$ & $+3.5$ &
$\textbf{+13.3}$ & $\textbf{+8.6}$ &
$\textbf{+12.6}$ &
$\textbf{+9.1}$ &
$+7.1$ & $+5.8$ &
$\textbf{+5.9}$ & $\underline{+7.5}$ &
$+6.6$ \\

TAG &
$-5.6$ & $-6.6$ &
$-0.2$ & $-4.2$ &
$-5.1$ &
$-4.3$ &
$-0.5$ & $-1.8$ &
$+3.5$ & $+4.1$ &
$+1.3$ \\

WSN & 
$+2.4$ & $+0.5$ &
$+4.3$ & $+3.3$ &
$+2.9$ & 
$+2.7$ & 
$+9.3$ & $+5.3$ &
$+4.6$ & $+6.6$ &
$+6.4$ \\

EWC &
$+0.4$ & $-2.5$ & 
$-3.5$ & $-5.6$ &
$-1.2$ &
$-2.5$ &
$+7.8$ & $+5.8$ &
$+4.7$ & $+6.7$ &
$+6.3$ \\

\hdashline

EWC-GI &
$-1.5$ & $+2.3$ &
$+6.3$ & $+6.8$ &
$\underline{+12.1}$ &
$+5.2$ &
$+8.8$ & $+5.7$ &
$+4.2$ & $+6.0$ &
$+6.2$ \\

SPG-FI &
$+3.2$ & $+1.0$ & 
$+1.5$ & $+4.4$ & 
$+6.6$ & 
$+3.4$ &
$\underline{+9.6}$ & $\textbf{+6.5}$ & 
$\underline{+5.4}$ & $+7.4$ & 
$\underline{+7.2}$ \\

\midrule

\textbf{SPG} &
$\underline{+5.5}$ & $\underline{+5.0}$ & 
$\underline{+8.9}$ & $\underline{+7.9}$ & 
$+10.2$ & 
$\underline{+7.5}$ &
$\textbf{+9.8}$ & $+5.6$ & 
$\textbf{+5.9}$ & $\textbf{+7.8}$ & 
$\textbf{+7.3}$ \\

\bottomrule
\end{tabular}
}
\vspace{-5mm}
\end{table}

\begin{table}[t]
\centering
\caption{
Backward transfer results in percent. 
}
\label{tab:results_bwt}
\resizebox{\hsize}{!}{
\begin{tabular}{crrrrrrrrrrr}
\toprule
& \multicolumn{6}{c}{Dissimilar tasks} & \multicolumn{5}{c}{Similar tasks} \\ \cmidrule(lr){2-7} \cmidrule(lr){8-12}
Model & 
\multicolumn{1}{c}{C-10} & \multicolumn{1}{c}{C-20} & \multicolumn{1}{c}{T-10} & \multicolumn{1}{c}{T-20} & \multicolumn{1}{c}{I-100} & \multicolumn{1}{c}{(Avg.)} &
\multicolumn{1}{c}{FC-10} & \multicolumn{1}{c}{FC-20} & \multicolumn{1}{c}{FE-10} & \multicolumn{1}{c}{FE-20} & \multicolumn{1}{c}{(Avg.)} \\
\midrule

NCL &
$-9.9$ & $-20.5$ &
$-3.4$ & $-9.6$ & 
$-17.9$ &
$-12.3$ &
$+2.1$ & $-0.9$ &
$+1.2$ & $\underline{+0.9}$ &
$+0.8$ \\

A-GEM &
$-14.3$ & $-21.8$ & 
$-4.0$ & $-9.7$ & 
$-16.6$ &
$-13.3$ &
$+0.2$ & $-1.0$ &
$\underline{+1.5}$ & $\underline{+0.9}$ & 
$+0.4$ \\

PGN &
$\textbf{0.0}$ & $\textbf{0.0}$ & 
$\textbf{0.0}$ & $\textbf{0.0}$ & 
$\underline{0.0}$ & 
$\textbf{0.0}$ &
$0.0$ & $0.0$ & 
$0.0$ & $0.0$ &
$0.0$ \\

PathNet &
$\textbf{0.0}$ & $\textbf{0.0}$ & 
$\textbf{0.0}$ & $\textbf{0.0}$ & 
$\underline{0.0}$ &
$\textbf{0.0}$ &
$0.0$ & $0.0$ &
$0.0$ & $0.0$ &
$0.0$ \\

HAT &
$\textbf{0.0}$ & $\textbf{0.0}$ &
$\textbf{0.0}$ & $\textbf{0.0}$ &
$\underline{0.0}$ &
$\textbf{0.0}$ &
$0.0$ & $0.0$ &
$0.0$ & $0.0$ & 
$0.0$ \\

CAT &
$\textbf{0.0}$ & $\textbf{0.0}$ & 
$\textbf{0.0}$ & $\textbf{0.0}$ &
N/A &
N/A &
$0.0$ & $0.0$ &
$0.0$ & $0.0$ & 
$0.0$ \\

SupSup &
$\textbf{0.0}$ & $\textbf{0.0}$ & 
$\textbf{0.0}$ & $\textbf{0.0}$ &
$\underline{0.0}$ & 
$\textbf{0.0}$ &
$0.0$ & $0.0$ &
$0.0$ & $0.0$ &
$0.0$ \\

UCL &
$-6.6$ & $-9.0$ &
$-3.4$ & $-7.7$ &
$-14.5$ &
$-8.2$ &
$\underline{+3.7}$ & $+2.2$ &
$+1.2$ & $+0.7$ & 
$\underline{+2.0}$ \\

SI &
$-12.8$ & $-5.7$ &
$-12.1$ & $-11.0$ &
$-17.6$ &
$-11.8$ &
$\textbf{+5.2}$ & $\underline{+2.3}$ &
$+0.9$ & $+0.7$ & 
$\textbf{+2.3}$ \\

TAG &
$\underline{-0.8}$ & $\underline{-1.6}$ &
$\underline{-0.4}$ & $\underline{-0.8}$ &
$\textbf{+0.9}$ &
$\underline{-0.5}$ &
$+0.4$ & $+0.4$ &
$-0.3$ & $0.0$ & 
$+0.1$ \\

WSN & 
$\textbf{0.0}$ & $\textbf{0.0}$ &
$\textbf{0.0}$ & $\textbf{0.0}$ &
$\underline{0.0}$ & 
$\textbf{0.0}$ & 
$0.0$ & $0.0$ &
$0.0$ & $0.0$ &
$0.0$ \\

EWC & 
$-6.4$ & $-13.9$ & 
$-3.9$ & $-7.8$ & 
$-22.7$ &
$-10.9$ &
$-1.3$ & $+1.6$ &
$+1.4$ & $+0.4$ & 
$+0.5$ \\

\hdashline

EWC-GI &
$-1.4$ & $-19.6$ &
$-1.6$ & $-13.4$ &
$-8.6$ &
$-8.9$ &
$0.0$ & $+0.1$ &
$\textbf{+1.6}$ & $\textbf{+1.7}$ &
$+0.9$ \\

SPG-FI &
$-10.7$ & $-10.3$ &
$-1.3$ & $-8.1$ &
$-7.0$ & 
$-7.5$ &
$+2.8$ & $+1.0$ &
$+1.4$ & $+0.8$ &
$+1.5$ \\

\midrule

\textbf{SPG} &
$-5.3$ & $-5.9$ &
$-4.4$ & $-3.4$ &
$-1.2$ &
$-4.0$ &
$+2.9$ & $\textbf{+2.8}$ &
$+0.9$ & $+0.5$ &
$+1.8$ \\

\bottomrule
\end{tabular}
}
\vspace{-3mm}
\end{table}

\begin{table}[h]
\centering
\caption{
Each cell reports how many percentage of parameters are completely blocked (i.e., importance of 1) just after learning task $t$. $T$ is the total number of tasks (e.g., $T=10$ for C-10).
}
\label{tab:blocking_ratio}
\resizebox{\columnwidth}{!}{
\begin{tabular}{ccrrrrrrrrr}
\toprule
& & \multicolumn{5}{c}{Dissimilar tasks} & \multicolumn{4}{c}{Similar tasks} \\ \cmidrule(lr){3-7} \cmidrule(lr){8-11}
$t$ & Model &
\multicolumn{1}{c}{C-10} & \multicolumn{1}{c}{C-20} & \multicolumn{1}{c}{T-10} & \multicolumn{1}{c}{T-20} & \multicolumn{1}{c}{I-100} & \multicolumn{1}{c}{FC-10} & \multicolumn{1}{c}{FC-20} & \multicolumn{1}{c}{FE-10} & \multicolumn{1}{c}{FE-20}  \\
\midrule

\multirow{2}{*}{$1$}
& HAT & 
$1.9\std{0.5}$ & $15.9\std{1.1}$ & 
$2.9\std{0.2}$ & $0.2\std{0.1}$ &
$23.5\std{0.6}$ & 
$0.3\std{0.0}$ & $6.3\std{1.0}$ &
$23.9\std{1.0}$ & $19.0\std{1.1}$ \\
& \textbf{SPG} & 
$0.1\std{0.0}$ & $0.2\std{0.0}$ &
$0.1\std{0.0}$ & $0.1\std{0.0}$ &
$0.1\std{0.0}$ &
$0.1\std{0.0}$ & $0.2\std{0.0}$ &
$0.1\std{0.0}$ & $0.1\std{0.0}$ \\

\hdashline

\multirow{2}{*}{$T/2$}
& HAT & 
$22.4\std{1.1}$ & $98.2\std{0.3}$ &
$36.7\std{1.9}$ & $21.1\std{1.4}$ &
$99.8\std{0.0}$ &
$3.4\std{0.3}$ & $65.7\std{2.4}$ &
$86.8\std{0.8}$ & $94.0\std{1.6}$ \\
& \textbf{SPG} & 
$0.9\std{0.1}$ & $2.9\std{0.3}$ &
$0.5\std{0.1}$ & $1.4\std{0.2}$ &
$4.1\std{0.4}$ &
$0.8\std{0.0}$ & $1.3\std{0.1}$ &
$0.6\std{0.3}$ & $1.0\std{0.2}$ \\

\hdashline

\multirow{2}{*}{$T$}
& HAT & 
$41.8\std{1.3}$ & $99.6\std{0.1}$ &
$57.5\std{1.8}$ & $38.6\std{2.1}$ &
$99.9\std{0.1}$ & 
$9.8\std{1.1}$ & $79.5\std{0.9}$ &
$97.9\std{0.3}$ & $97.5\std{0.9}$ \\
& \textbf{SPG} &
$2.0\std{0.2}$ & $5.2\std{0.6}$ &
$0.9\std{0.1}$ & $3.1\std{0.2}$ &
$6.3\std{0.6}$ &
$1.4\std{0.1}$ & $2.0\std{0.3}$ &
$1.1\std{0.5}$ & $1.5\std{0.3}$ \\

\bottomrule
\end{tabular}
}
\vspace{-2mm}
\end{table}

\begin{table*}[t]
\centering
\caption{
The results for pruning parameters based on the gradient-based importance.
}
\label{tab:app_pruning}
\resizebox{.95\hsize}{!}{
 \begin{tabular}{rrrrrrrrrr}
\toprule
& \multicolumn{5}{c}{Dissimilar tasks} & \multicolumn{4}{c}{Similar tasks} \\ \cmidrule(lr){2-6} \cmidrule(lr){7-10}
\multicolumn{1}{c}{Pruning strategy} & \multicolumn{1}{c}{C-10} & \multicolumn{1}{c}{C-20} & \multicolumn{1}{c}{T-10} & \multicolumn{1}{c}{T-20} & \multicolumn{1}{c}{I-100} &
 \multicolumn{1}{c}{FC-10} & \multicolumn{1}{c}{FC-20} & \multicolumn{1}{c}{FE-10} & \multicolumn{1}{c}{FE-20} \\
\midrule

Nothing & 
$75.5\std{1.0}$ & $78.9\std{1.9}$ &
$46.7\std{0.8}$ & $49.2\std{1.2}$ &
$48.8\std{1.2}$ & 
$74.5\std{3.0}$ & $87.8\std{2.6}$ &
$86.0\std{0.5}$ & $85.4\std{0.8}$ \\

\hdashline

Lowest $10$\% &
$73.6\std{2.4}$ & $75.4\std{4.8}$ &
$41.2\std{0.9}$ & $45.8\std{1.5}$ &
$43.3\std{2.6}$ & 
$74.0\std{2.9}$ & $85.5\std{1.9}$ &
$83.1\std{1.7}$ & $84.3\std{0.4}$ \\

Random $10$\% &
$69.5\std{2.1}$ & $70.5\std{4.2}$ &
$34.7\std{1.1}$ & $44.8\std{1.6}$ &
$40.8\std{1.2}$ & 
$71.3\std{2.0}$ & $84.3\std{3.7}$ &
$70.8\std{4.6}$ & $72.3\std{3.1}$ \\

Highest $10$\% &
$19.4\std{8.5}$ & $24.6\std{8.1}$ &
$6.2\std{1.1}$ & $13.2\std{2.4}$ &
$17.1\std{3.6}$ & 
$39.0\std{0.6}$ & $49.3\std{4.1}$ &
$55.5\std{9.7}$ & $19.8\std{8.4}$ \\

\hdashline

Lowest $20$\% &
$68.6\std{5.1}$ & $72.7\std{6.7}$ &
$36.4\std{3.3}$ & $43.1\std{1.8}$ &
$39.4\std{4.1}$ & 
$73.3\std{3.9}$ & $85.5\std{3.7}$ &
$61.5\std{2.7}$ & $81.6\std{3.5}$ \\

Random $20$\% &
$58.3\std{2.2}$ & $58.7\std{7.4}$ &
$20.4\std{0.6}$ & $37.2\std{3.1}$ &
$32.3\std{2.6}$ & 
$68.5\std{3.0}$ & $85.0\std{3.6}$ &
$37.2\std{3.4}$ & $43.5\std{5.4}$ \\

Highest $20$\% &
$10.8\std{1.0}$ & $22.5\std{5.6}$ &
$5.5\std{0.6}$ & $10.2\std{0.2}$ &
$11.7\std{1.3}$ & 
$38.8\std{0.0}$ & $49.3\std{4.1}$ &
$20.7\std{5.8}$ & $7.8\std{1.5}$ \\

\bottomrule
\end{tabular}
}

\vspace{-3mm}

\end{table*}

\subsubsection{Validity of Gradient-based Importance}

We further analyze how the proposed gradient-based importance metric
indicates the contribution of parameters with different importance values to the performance of each task. Specifically, we evaluate the accuracy on the first task of each dataset (no continual learning) 
after pruning some parameters based on their importance so that we can confirm whether the importance metric is co-related with the performance change.

The following four strategies for choosing which parameters to prune are compared --
\textbf{(1) Nothing}: 
we do not prune any parameters. 
\textbf{(2) Random $\bm{n}$\%}: 
$n\%$ of parameters are randomly pruned regardless of their importance.
\textbf{(3) Lowest $\bm{n}$\%}: 
the parameters with the lowest $n\%$ of importance are pruned.
\textbf{(4) Highest $\bm{n}$\%}:
the parameters with the highest $n\%$ of importance are pruned.

The average results over 5 different seeds are presented in \cref{tab:app_pruning}.
It can be clearly observed that pruning parameters with higher importance degrades the performance more (e.g., ``Lowest $10$\%'' is much better than ``Highest $10$\%''). When it is ``Highest 20\%'', the accuracy is almost like random chance classification (e.g., it is 10.8\% for C-10 while the random chance also gives 10\%). We believe this observation implies that our gradient-based importance metric effectively indicates the importance of parameters.

\begin{figure}[t]

\begin{minipage}{.48\hsize}
\centering
\includegraphics[width=\hsize]{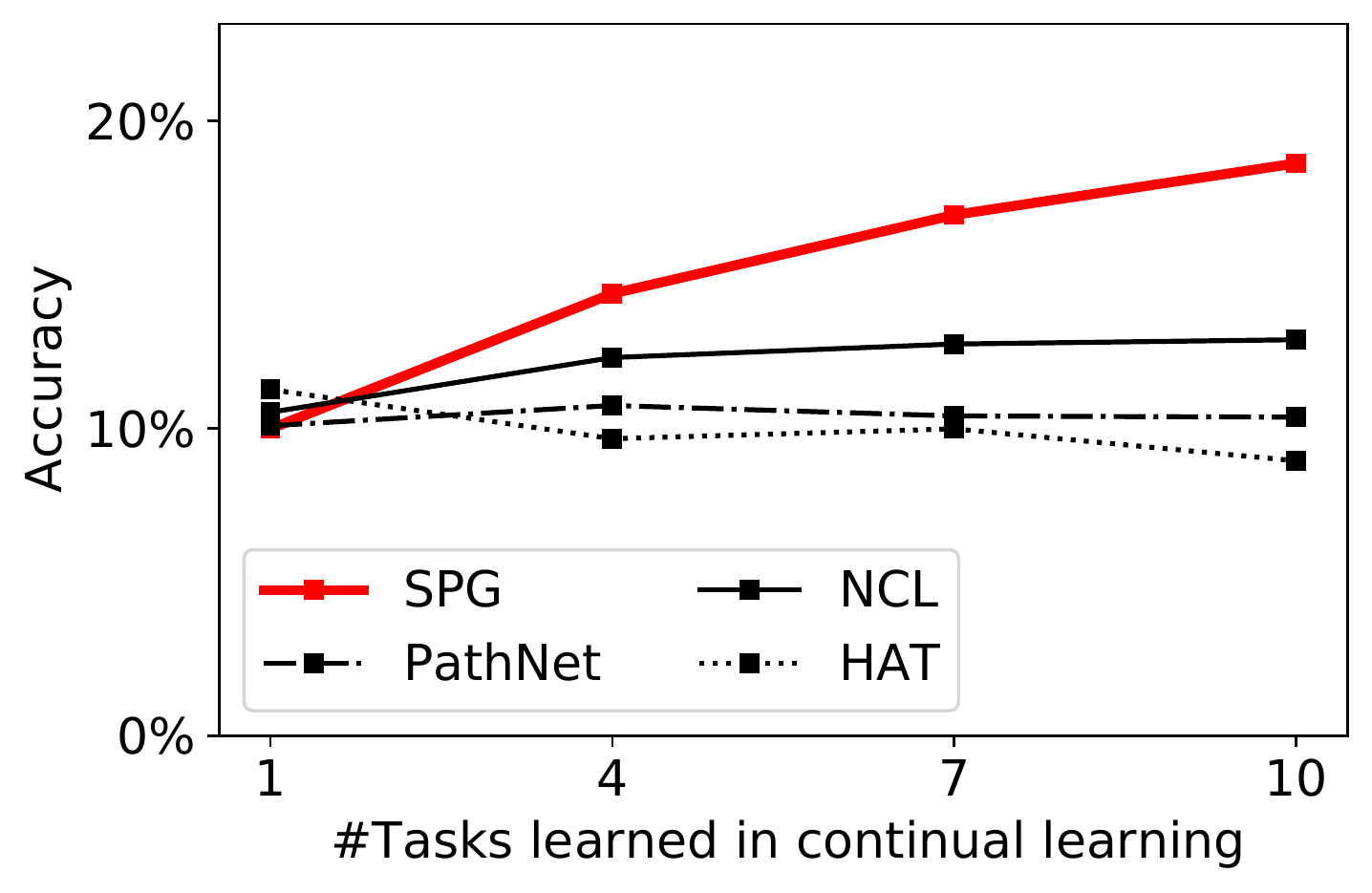}
\subcaption{Fine-tuning for TinyImageNet after CL for C-10}
\end{minipage}
\hfill
\begin{minipage}{.48\hsize}
\centering
\includegraphics[width=\hsize]{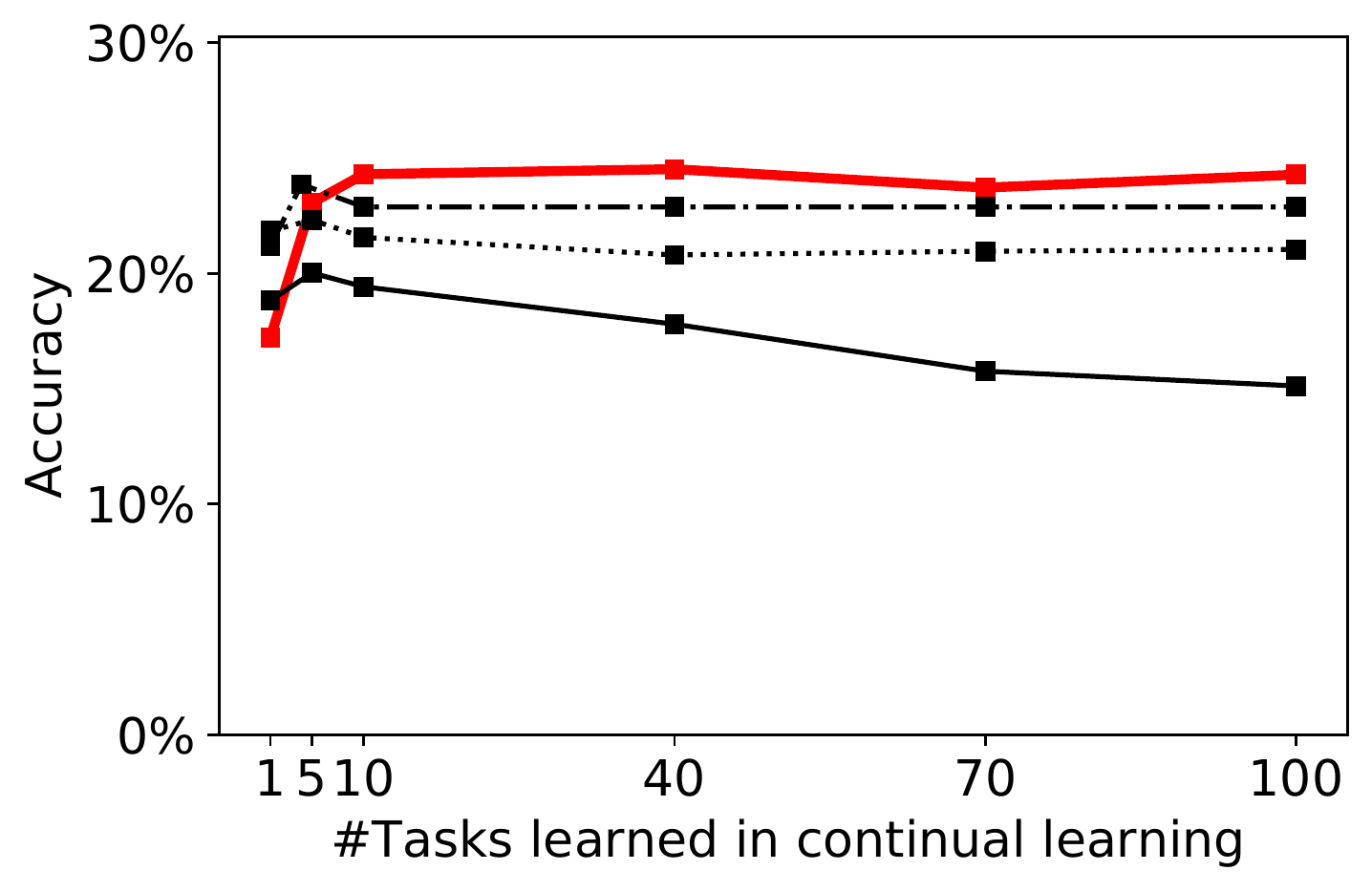}
\subcaption{Fine-tuning for CIFAR100 after CL for I-100}
\end{minipage}

\caption{
The learning of representation through continual learning.
The x-axis means the number of tasks learned in continual learning (CL).
The pair of a CL/non-CL dataset for (a) and (b) is C-10/TinyImageNet and I-100/CIFAR100, respectively.
}
\label{fig:cl_representation}

\vspace{-5mm}

\end{figure}

\begin{table*}[tb]
\centering
\caption{
The results for the ablation studies.
}
\label{tab:results_ablation}
\resizebox{.95\hsize}{!}{
\begin{tabular}{lrrrrrcrrrrc}
\toprule
& \multicolumn{6}{c}{Dissimilar tasks} & \multicolumn{5}{c}{Similar tasks} \\ \cmidrule(lr){2-7} \cmidrule(lr){8-12}
\multicolumn{1}{c}{Ablation} & \multicolumn{1}{c}{C-10} & \multicolumn{1}{c}{C-20} & \multicolumn{1}{c}{T-10} & \multicolumn{1}{c}{T-20} & \multicolumn{1}{c}{I-100} & \textbf{(Avg.)} & \multicolumn{1}{c}{FC-10} & \multicolumn{1}{c}{FC-20} & \multicolumn{1}{c}{FE-10} & \multicolumn{1}{c}{FE-20} & \textbf{(Avg.)} \\
\midrule

\textbf{SPG} & 
$\textbf{67.7}\std{0.3}$ & $\textbf{75.9}\std{1.1}$ &
$\textbf{48.4}\std{0.3}$ & $\textbf{59.1}\std{0.5}$ &
$\textbf{58.1}\std{0.5}$ & 
$\textbf{61.8}$ & 
$\textbf{87.0}\std{0.9}$ & $\textbf{87.1}\std{0.2}$ &
$\underline{87.7}\std{0.2}$ & $87.9\std{0.1}$ &
$\textbf{87.4}$ \\

\hdashline

SPG (w/o CHI) & 
$65.4\std{0.6}$ & $70.4\std{2.2}$ &
$46.9\std{0.7}$ & $55.3\std{0.8}$ &
$\underline{57.6}\std{0.4}$ & 
$59.1$ & 
$\underline{86.2}\std{0.3}$ & $\underline{86.9}\std{0.5}$ &
$\textbf{87.9}\std{0.3}$ & $\textbf{88.1}\std{0.3}$ &
$\underline{87.3}$ \\

SPG (w/o SMH) &
$\underline{66.1}\std{1.0}$ & $\underline{74.8}\std{0.3}$ &
$\underline{47.2}\std{0.5}$ & $\underline{56.9}\std{0.4}$ &
$55.2\std{0.8}$ & 
$\underline{60.0}$ & 
$85.8\std{0.7}$ & $86.8\std{0.5}$ &
$87.2\std{0.2}$ & $87.5\std{0.2}$ &
$86.8$ \\

SPG (w/ hard-masking) &
$63.6\std{0.4}$ & $73.2\std{1.2}$ &
$46.8\std{0.4}$ & $53.7\std{0.2}$ &
$51.1\std{0.3}$ &
$57.7$ & 
$84.3\std{0.4}$ & $86.3\std{0.5}$ &
$86.8\std{0.0}$ & $\underline{88.0}\std{0.2}$ &
$86.3$ \\

\bottomrule
\end{tabular}
}

\vspace{-3mm}

\end{table*}

\subsubsection{Better Representation Learning of SPG}

We found that SPG's stronger performance is manifested in its strong representations learning. 
We conduct additional experiments from the perspective of  representation learning.
In particular, a model that has just incrementally learned some tasks of a CL dataset (e.g., 5 tasks of C-10) are used as a frozen feature extractor to learn another dataset (not split into tasks), which we call ``non-CL dataset'', by fine-tuning a new classifier using the dataset, e.g., CIFAR100 or TinyImageNet (a non-CL dataset contains all classes of the dataset shown in \cref{tab:statisctics_datasets}). 
We evaluate the test accuracy for a non-CL dataset. Three pairs of a CL/non-CL dataset are tested: (1) C-10/TinyImageNet, (2) I-100/CIFAR100, and (3) T-10/CIFAR100. We here only show the results for (1) and (2) due to the space limitations (all the results are provided in \cref{sec:app_representation}).

\cref{fig:cl_representation}(a) shows that SPG learns better representations in continual learning than baselines (we use NCL and two strong performing baselines from \cref{tab:results_acc}). HAT even deteriorates, which indicates that hard-masking of some parameters/units leave less network capacity to learn good features. 
\cref{fig:cl_representation}(b) shows that while NCL significantly degrades due to its serious forgetting in such an extreme continual learning on I-100 (with 100 tasks), SPG keeps the best after learning a few tasks.
These clearly confirm that SPG learns better representations to enable better continual learning.

\subsection{Ablation Studies} \label{sec:ablation}

As SPG has three core mechanisms that contribute to its performance, we evaluate whether each of them is beneficial.
The first one is \textit{\underline{c}ross-\underline{h}ead \underline{i}mportance} (CHI), which is introduced in \cref{eq:gamma_i_t} with the motivation of suppressing the update of important parameters for previous tasks more. In the ablation \textbf{SPG (w/o CHI)}, $\bm{\gamma}_i^t$ is replaced with $\bm{\gamma}_i^{t,t}$ in \cref{eq:gamma_i_t} and the previous tasks' heads are not used for computing importance.
The second one is \underline{\textit{s}}oft-\underline{\textit{m}}asking of each \underline{\textit{h}}ead (SMH), which is introduced in \cref{eq:g_ht_prime} for the purpose of balancing the training of the feature extractor and the classification head. In the ablation \textbf{SPG (w/o SMH)}, \cref{eq:g_ht_prime} is not employed.
The last one is soft-masking (not hard-masking), which is the core technique of SPG to keep most parameters trainable while mitigating CF. While the reported results so far are based on soft-masking, it is also possible to convert the importance to binary (0 or 1) masks using a pre-defined threshold. In the ablation \textbf{SPG (w/ hard-masking)}, if the threshold is 0.6, this variant treats importance larger than 0.6 as 1 (blocking updates of parameters), otherwise 0 (not blocking). We search for the best threshold for each dataset from $\{0.2,0.4,0.6,0.8\}$.

The ablation results are presented in \cref{tab:results_ablation}.
For CHI, it improves the performance especially in dissimilar tasks (up to 5.5\% for C-20) by blocking more gradient flow to mitigate CF. On the other hand, CHI does not contributes much for similar tasks, which is reasonable as blocking parameters becomes less important in similar tasks. Nevertheless, 
SPG with CHI still works the best for similar tasks on average and CHI does not cause side effects. More quantitative analysis of how CHI contributes to suppressing parameter updates is given in \cref{sec:app_chi}.
For SMH, it delivers positive performance gains both in dissimilar and similar tasks. These results indicate that the lack of balance in the training between the feature extractor and head, which can happen without SMH, adversely affects the performance. The promotion of knowledge transfer by SMH becomes most prominent for I-100 (up to 2.9\%) as the effectiveness of knowledge transfer is more important in such an extreme case with more tasks.
For whether masking should be soft or hard, SPG with hard-masking is significantly worse than SPG with soft-masking, which demonstrates that it is difficult to find a good threshold for hard-masking. Soft-masking is more flexible and effective.
These results clearly confirm that SPG enjoys all of the three different mechanisms.

\section{Conclusion} \label{sec:conclusion}

To overcome the difficulty of balancing forgetting prevention and knowledge transfer in continual learning, we proposed a novel and simple method, called SPG, that blocks/masks parameters not completely but partially to give the model more flexibility and capacity to learn. The proposed soft-masking mechanism not only overcomes CF but also performs knowledge transfer automatically. Although it is conceptually related to the regularization approach, as we have argued and evaluated, it markedly outperforms the regularization approach. 
Extensive experiments demonstrate that SPG markedly outperforms all the strong baselines. 

\section*{Acknowledgements}
The work of Zixuan Ke, Gyuhak Kim, and Bing Liu was supported in part by a research contract from KDDI Research, Inc. and three NSF grants (IIS-1910424, IIS-1838770, and CNS-2225427).


\bibliography{icml2023_paper}
\bibliographystyle{icml2023}

\newpage
\appendix
\onecolumn

\label{sec:appendix}

\section{Forward Transfer} \label{sec:app_results_fwt}

\begin{figure}[h]

\centering
\begin{minipage}{.27\hsize}
\centering
\includegraphics[width=\hsize]{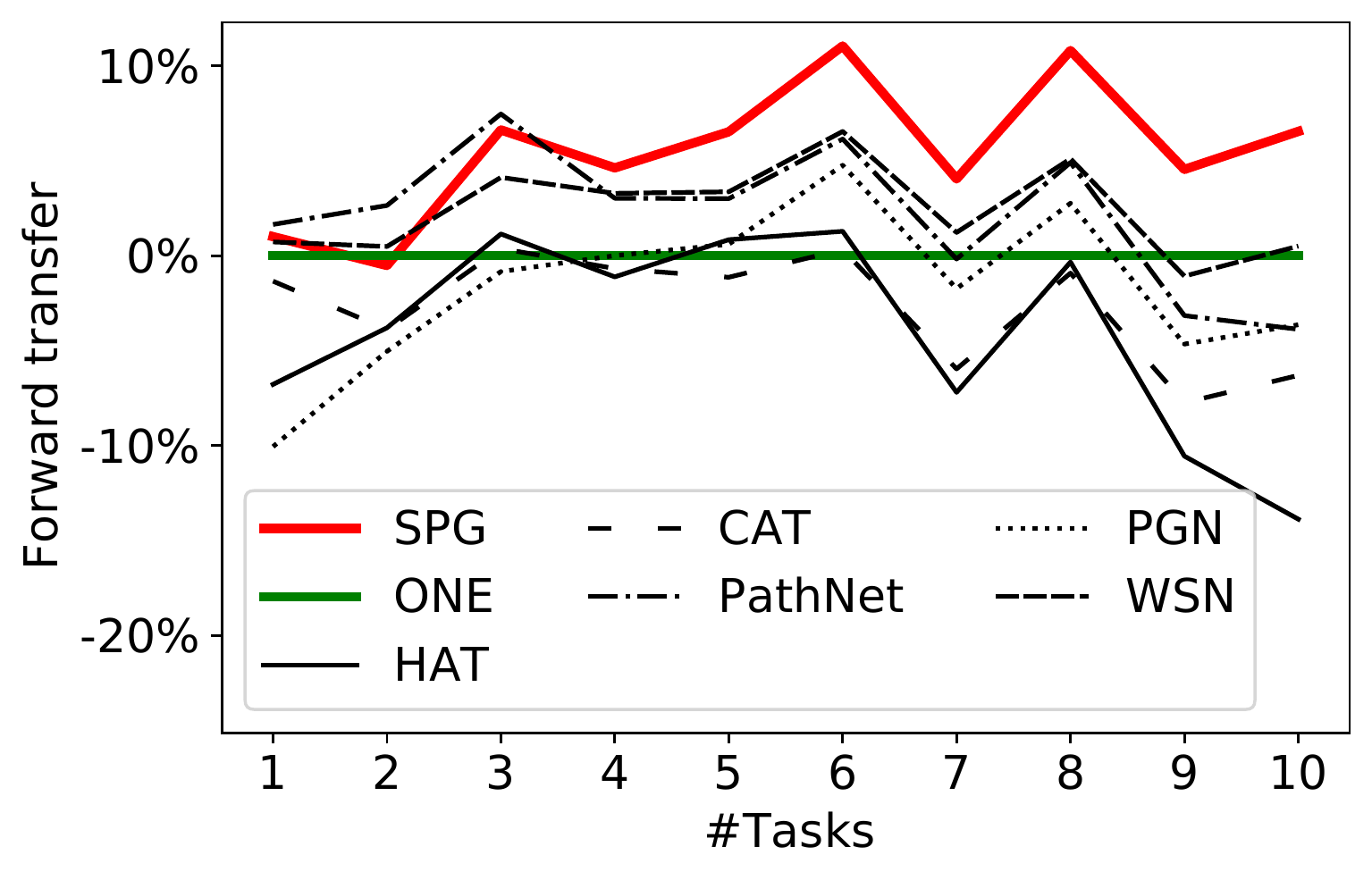}
\subcaption{C-10}
\end{minipage}
\begin{minipage}{.27\hsize}
\centering
\includegraphics[width=\hsize]{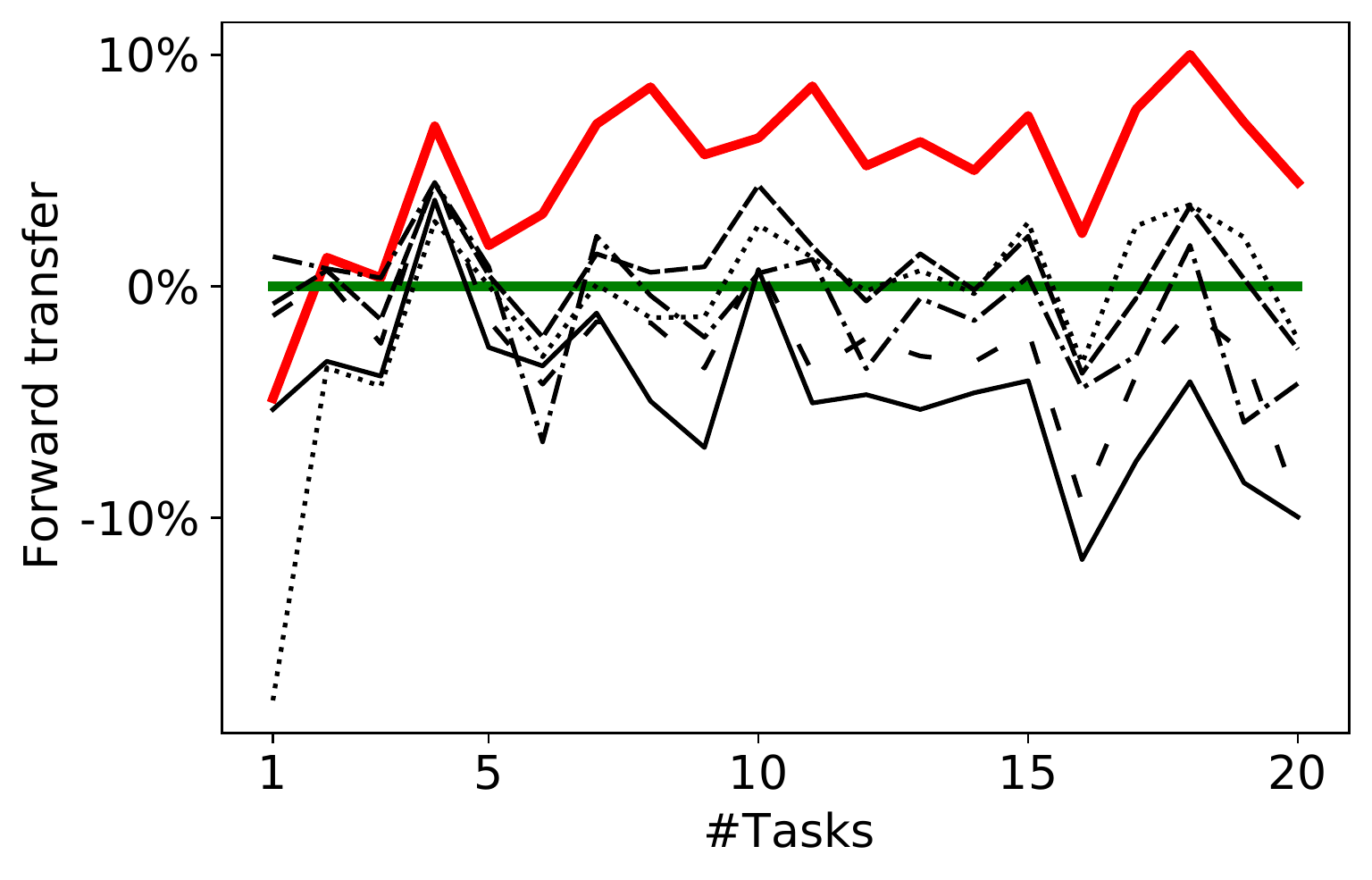}
\subcaption{C-20}
\end{minipage}
\begin{minipage}{.27\hsize}
\centering
\includegraphics[width=\hsize]{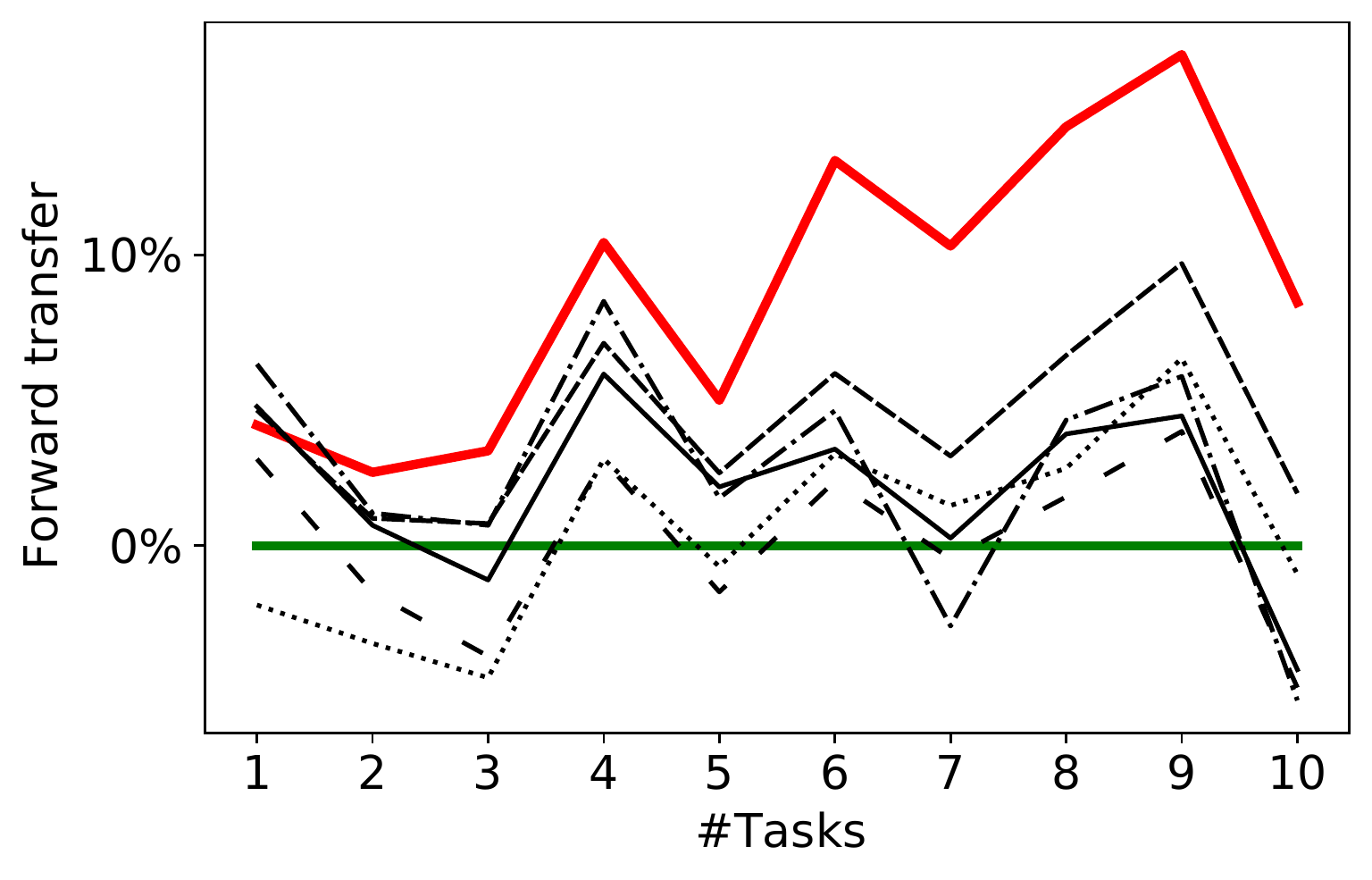}
\subcaption{T-10}
\end{minipage}

\medskip

\begin{minipage}{.27\hsize}
\centering
\includegraphics[width=\hsize]{images/fwt_T-20.pdf}
\subcaption{T-20}
\end{minipage}
\begin{minipage}{.27\hsize}
\centering
\includegraphics[width=\hsize]{images/fwt_I-100.pdf}
\subcaption{I-100}
\end{minipage}
\begin{minipage}{.27\hsize}
\centering
\includegraphics[width=\hsize]{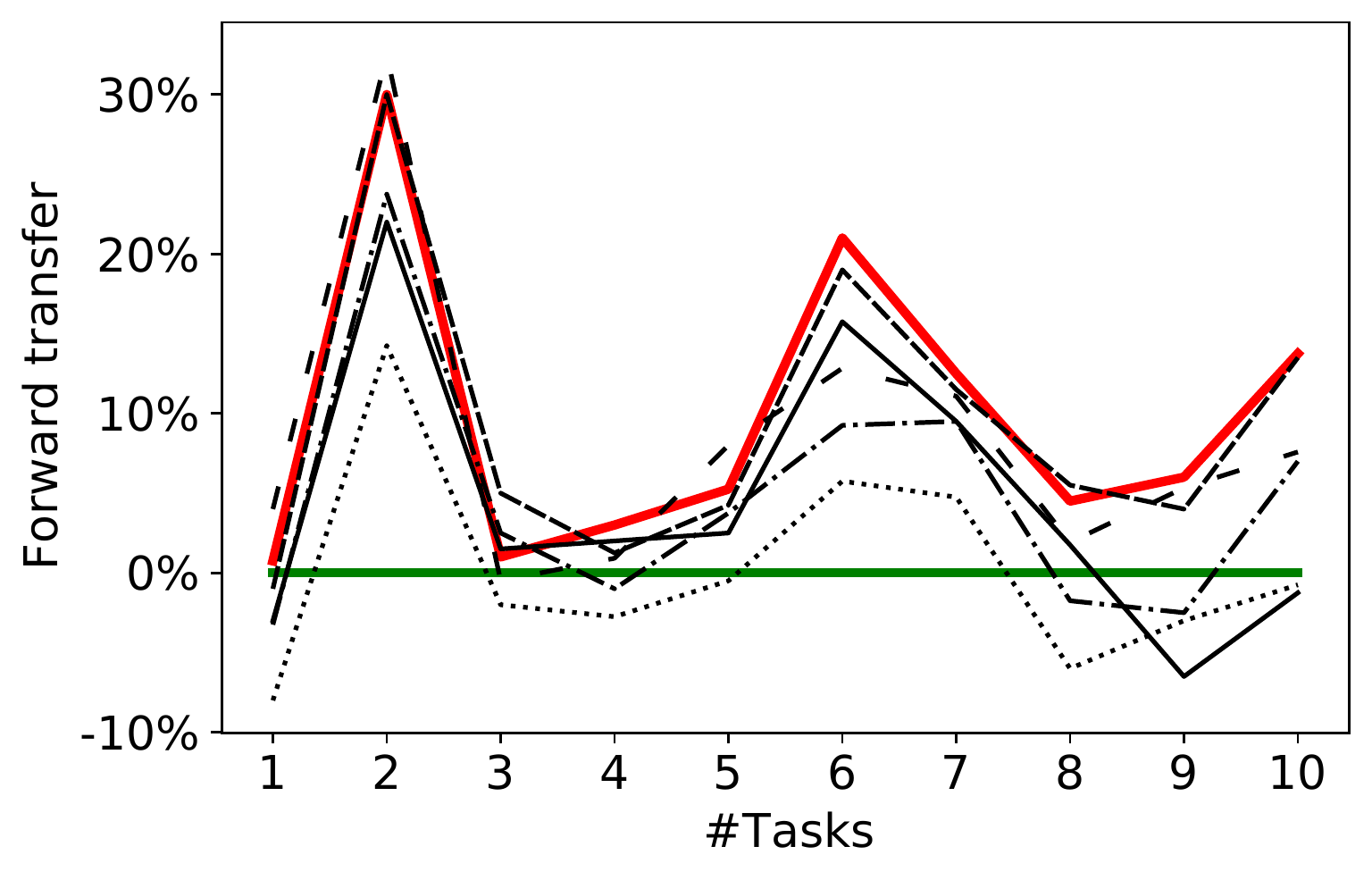}
\subcaption{FC-10}
\end{minipage}

\medskip

\begin{minipage}{.27\hsize}
\centering
\includegraphics[width=\hsize]{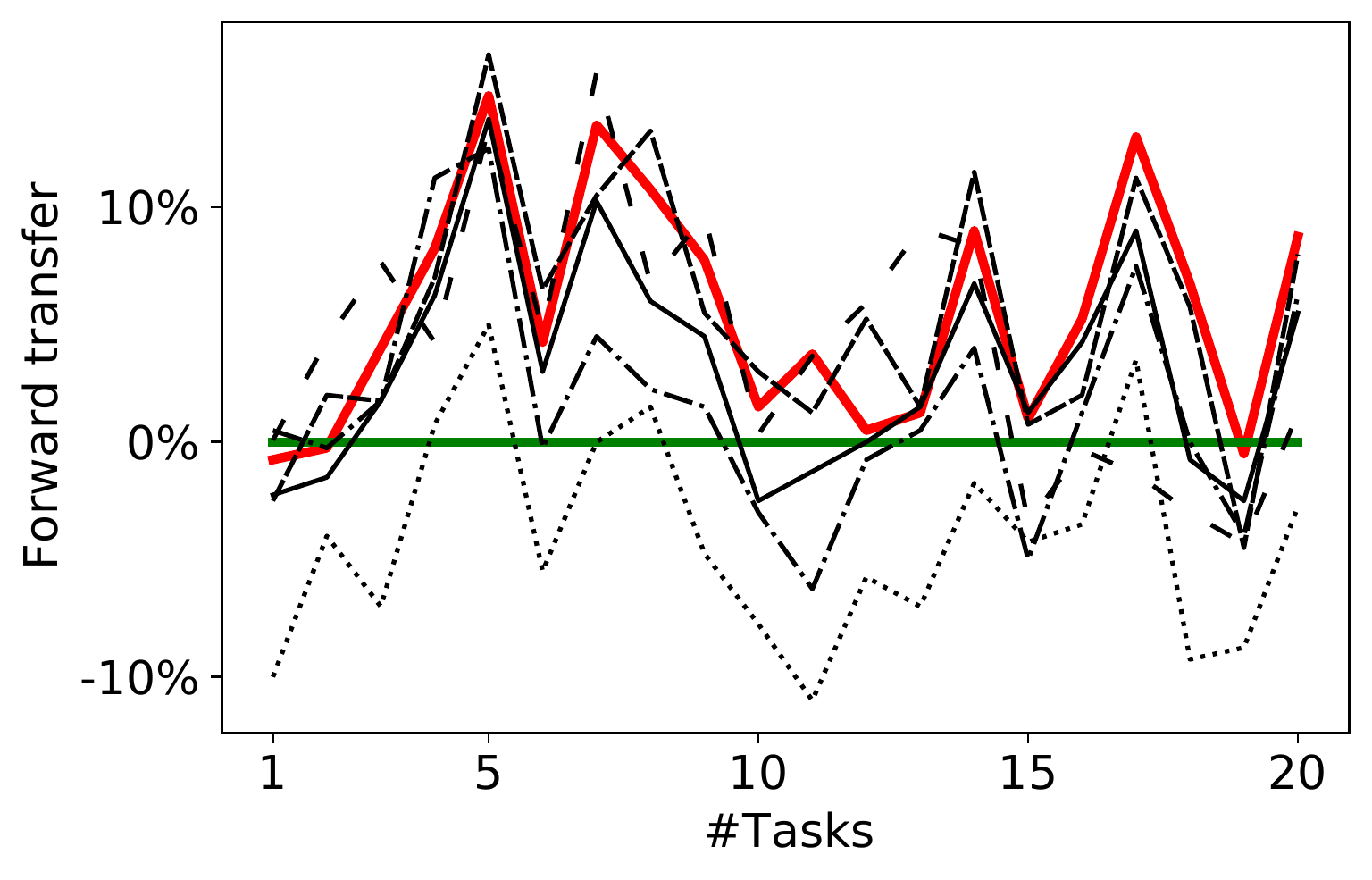}
\subcaption{FC-20}
\end{minipage}
\begin{minipage}{.27\hsize}
\centering
\includegraphics[width=\hsize]{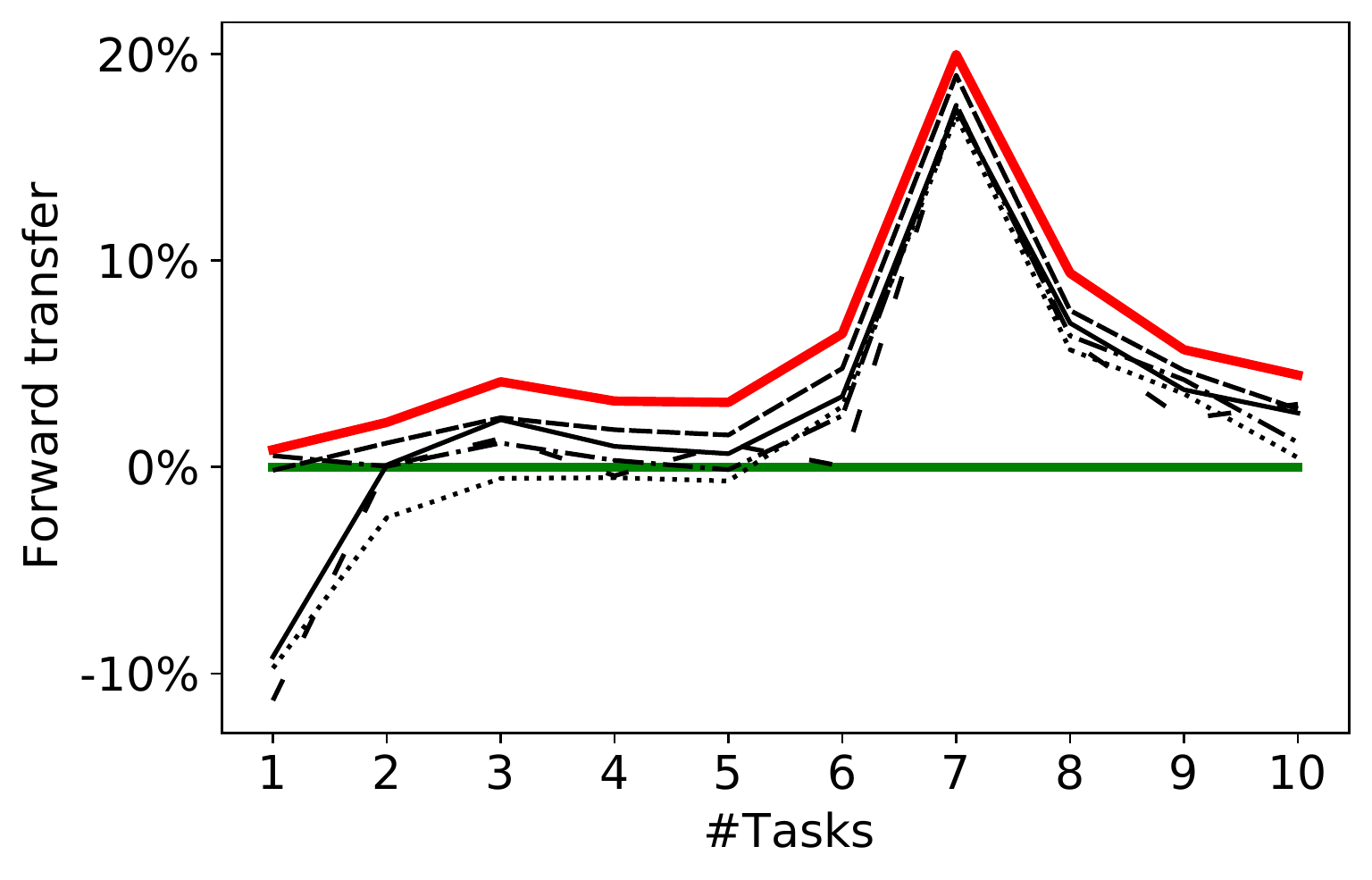}
\subcaption{FE-10}
\end{minipage}
\begin{minipage}{.27\hsize}
\centering
\includegraphics[width=\hsize]{images/fwt_FE-20.pdf}
\subcaption{FE-20}
\end{minipage}

\caption{Forward transfer results. (a) to (e) are for dissimilar tasks, and (f) to (i) are for similar tasks.}
\label{fig:app_results_fwt}

\end{figure}

\cref{fig:results_fwt} in the main paper shows the forward transfer plots for some datasets but not all due to space limitations. \cref{fig:app_results_fwt} presents the forward transfer results for all datasets. 
It can be clearly seen that SPG has the best positive forward transfer and keeps or even grows it constantly in all cases. On the other hand, the other parameter isolation-based methods, PGN, PathNet, HAT, CAT, and WSN lose their ability for the forward transfer in later tasks for the dissimilar task (i.e., (a) to (e)).

\section{Capacity Consumption at Each Layer} \label{sec:app_capacity_layerwise}

We show the percentage of parameters in the whole network that are fully blocked in \cref{tab:blocking_ratio} in the main text of the paper. Here \cref{tab:app_blocking_ratio_layerwise} presents the same result for each layer. We use AlexNet as the backbone, and it has three convolution layers followed by two fully-connected layers.

\begin{table}[p]
\centering
\caption{
The percentage of parameters that are completely blocked for each layer.
$T$ is the total number of tasks (e.g., $T = 10$ for C-10).
}
\label{tab:app_blocking_ratio_layerwise}

\vspace{2mm}

\begin{minipage}{.85\hsize}
\centering
\subcaption{Results for the 1st convolution layer.}
\resizebox{\columnwidth}{!}{
\begin{tabular}{ccrrrrrrrrr}
\toprule
& & \multicolumn{5}{c}{Dissimilar tasks} & \multicolumn{4}{c}{Similar tasks} \\ \cmidrule(lr){3-7} \cmidrule(lr){8-11}
$t$ & Model &
\multicolumn{1}{c}{C-10} & \multicolumn{1}{c}{C-20} & \multicolumn{1}{c}{T-10} & \multicolumn{1}{c}{T-20} & \multicolumn{1}{c}{I-100} &
\multicolumn{1}{c}{FC-10} & \multicolumn{1}{c}{FC-20} & \multicolumn{1}{c}{FE-10} & \multicolumn{1}{c}{FE-20}  \\
\midrule

\multirow{2}{*}{$1$}
& HAT & 
$25.0\std{5.3}$ & $24.4\std{8.4}$ & 
$31.6\std{6.0}$ & $14.4\std{5.5}$ &
$60.0\std{4.4}$ & 
$17.8\std{4.4}$ & $40.0\std{8.7}$ &
$75.0\std{3.1}$ & $67.2\std{9.9}$ \\
& \textbf{SPG} & 
$0.0\std{0.0}$ & $0.0\std{0.0}$ &
$0.0\std{0.0}$ & $0.0\std{0.0}$ &
$0.0\std{0.0}$ &
$0.0\std{0.0}$ & $0.0\std{0.0}$ &
$0.0\std{0.0}$ & $0.0\std{0.0}$ \\

\hdashline

\multirow{2}{*}{$T/2$}
& HAT & 
$77.8\std{2.7}$ & $98.4\std{1.7}$ &
$83.4\std{3.6}$ & $79.1\std{2.5}$ &
$100.0\std{0.0}$ &
$73.4\std{7.6}$ & $100.0\std{0.0}$ &
$98.8\std{1.2}$ & $99.4\std{0.8}$ \\
& \textbf{SPG} & 
$0.0\std{0.0}$ & $0.0\std{0.0}$ &
$0.0\std{0.0}$ & $0.0\std{0.0}$ &
$0.0\std{0.0}$ &
$0.0\std{0.0}$ & $0.0\std{0.0}$ &
$0.0\std{0.0}$ & $0.0\std{0.0}$ \\

\hdashline

\multirow{2}{*}{$T$}
& HAT & 
$95.6\std{1.5}$ & $100.0\std{0.0}$ &
$97.2\std{1.2}$ & $95.3\std{1.0}$ &
$100.0\std{0.0}$ & 
$98.8\std{1.8}$ & $100.0\std{0.0}$ &
$100.0\std{0.0}$ & $100.0\std{0.0}$ \\
& \textbf{SPG} & 
$0.0\std{0.0}$ & $0.0\std{0.0}$ &
$0.0\std{0.0}$ & $0.0\std{0.0}$ &
$0.0\std{0.0}$ &
$0.0\std{0.0}$ & $0.0\std{0.0}$ &
$0.0\std{0.0}$ & $0.0\std{0.0}$ \\

\bottomrule
\end{tabular}
}
\end{minipage}

\vspace{2mm}

\begin{minipage}{.85\hsize}
\centering
\subcaption{Results for the 2nd convolution layer.}
\resizebox{\columnwidth}{!}{
\begin{tabular}{ccrrrrrrrrr}
\toprule
& & \multicolumn{5}{c}{Dissimilar tasks} & \multicolumn{4}{c}{Similar tasks} \\ \cmidrule(lr){3-7} \cmidrule(lr){8-11}
$t$ & Model &
\multicolumn{1}{c}{C-10} & \multicolumn{1}{c}{C-20} & \multicolumn{1}{c}{T-10} & \multicolumn{1}{c}{T-20} & \multicolumn{1}{c}{I-100} &
\multicolumn{1}{c}{FC-10} & \multicolumn{1}{c}{FC-20} & \multicolumn{1}{c}{FE-10} & \multicolumn{1}{c}{FE-20}  \\
\midrule

\multirow{2}{*}{$1$}
& HAT & 
$5.6\std{1.6}$ & $7.3\std{2.9}$ & 
$8.9\std{2.3}$ & $1.5\std{1.1}$ &
$33.9\std{5.0}$ & 
$2.2\std{0.6}$ & $12.0\std{2.0}$ &
$55.7\std{6.6}$ & $45.7\std{12.1}$ \\
& \textbf{SPG} & 
$0.1\std{0.0}$ & $0.1\std{0.1}$ &
$0.0\std{0.0}$ & $0.1\std{0.0}$ &
$0.0\std{0.0}$ &
$0.0\std{0.0}$ & $0.0\std{0.0}$ &
$0.0\std{0.1}$ & $0.0\std{0.0}$ \\

\hdashline

\multirow{2}{*}{$T/2$}
& HAT & 
$52.4\std{2.5}$ & $98.0\std{1.7}$ &
$63.7\std{6.2}$ & $56.2\std{4.5}$ &
$100.0\std{0.0}$ &
$41.3\std{4.7}$ & $98.3\std{1.2}$ &
$98.4\std{1.0}$ & $99.2\std{1.0}$ \\
& \textbf{SPG} & 
$0.6\std{0.1}$ & $2.1\std{0.4}$ &
$0.3\std{0.1}$ & $0.9\std{0.3}$ &
$1.5\std{0.3}$ &
$0.1\std{0.1}$ & $0.2\std{0.1}$ &
$0.5\std{0.5}$ & $0.6\std{0.2}$ \\

\hdashline

\multirow{2}{*}{$T$}
& HAT & 
$81.0\std{2.0}$ & $99.8\std{0.3}$ &
$90.2\std{2.8}$ & $86.4\std{4.5}$ &
$100.0\std{0.0}$ & 
$86.9\std{2.2}$ & $99.4\std{0.3}$ &
$100.0\std{0.0}$ & $100.0\std{0.0}$ \\
& \textbf{SPG} & 
$1.1\std{0.2}$ & $3.1\std{0.5}$ &
$0.4\std{0.1}$ & $1.5\std{0.3}$ &
$1.8\std{0.4}$ &
$0.2\std{0.1}$ & $0.3\std{0.2}$ &
$0.8\std{0.7}$ & $0.9\std{0.3}$ \\

\bottomrule
\end{tabular}
}
\end{minipage}

\vspace{2mm}

\begin{minipage}{.85\hsize}
\centering
\subcaption{Results for the 3rd convolution layer.}
\resizebox{\columnwidth}{!}{
\begin{tabular}{ccrrrrrrrrr}
\toprule
& & \multicolumn{5}{c}{Dissimilar tasks} & \multicolumn{4}{c}{Similar tasks} \\ \cmidrule(lr){3-7} \cmidrule(lr){8-11}
$t$ & Model &
\multicolumn{1}{c}{C-10} & \multicolumn{1}{c}{C-20} & \multicolumn{1}{c}{T-10} & \multicolumn{1}{c}{T-20} & \multicolumn{1}{c}{I-100} &
\multicolumn{1}{c}{FC-10} & \multicolumn{1}{c}{FC-20} & \multicolumn{1}{c}{FE-10} & \multicolumn{1}{c}{FE-20}  \\
\midrule

\multirow{2}{*}{$1$}
& HAT & 
$4.8\std{1.3}$ & $8.7\std{1.7}$ & 
$7.5\std{1.1}$ & $0.7\std{0.3}$ &
$30.6\std{2.8}$ & 
$1.0\std{0.2}$ & $8.0\std{1.5}$ &
$43.7\std{6.5}$ & $36.4\std{6.2}$ \\
& \textbf{SPG} & 
$0.1\std{0.0}$ & $0.1\std{0.0}$ &
$0.1\std{0.0}$ & $0.1\std{0.0}$ &
$0.1\std{0.0}$ &
$0.1\std{0.0}$ & $0.2\std{0.1}$ &
$0.0\std{0.0}$ & $0.0\std{0.0}$ \\

\hdashline

\multirow{2}{*}{$T/2$}
& HAT & 
$42.6\std{3.6}$ & $98.7\std{0.7}$ &
$60.3\std{4.1}$ & $47.1\std{2.3}$ &
$100.0\std{0.0}$ &
$14.7\std{1.2}$ & $88.7\std{4.2}$ &
$98.0\std{0.7}$ & $99.1\std{0.4}$ \\
& \textbf{SPG} & 
$0.7\std{0.1}$ & $2.1\std{0.4}$ &
$0.3\std{0.1}$ & $1.0\std{0.2}$ &
$1.2\std{0.2}$ &
$0.5\std{0.1}$ & $1.0\std{0.3}$ &
$0.3\std{0.2}$ & $0.3\std{0.1}$ \\

\hdashline

\multirow{2}{*}{$T$}
& HAT & 
$66.8\std{2.7}$ & $99.7\std{0.5}$ &
$87.1\std{2.2}$ & $74.1\std{3.4}$ &
$100.0\std{0.0}$ & 
$43.1\std{4.0}$ & $95.3\std{1.9}$ &
$99.8\std{0.5}$ & $99.8\std{0.2}$ \\
& \textbf{SPG} & 
$1.2\std{0.2}$ & $3.1\std{0.7}$ &
$0.4\std{0.1}$ & $1.5\std{0.3}$ &
$1.8\std{0.2}$ &
$0.9\std{0.2}$ & $1.4\std{0.4}$ &
$0.5\std{0.3}$ & $0.5\std{0.1}$ \\

\bottomrule
\end{tabular}
}
\end{minipage}

\vspace{2mm}

\begin{minipage}{.85\hsize}
\centering
\subcaption{Results for the 4th fully-connected layer.}
\resizebox{\columnwidth}{!}{
\begin{tabular}{ccrrrrrrrrr}
\toprule
& & \multicolumn{5}{c}{Dissimilar tasks} & \multicolumn{4}{c}{Similar tasks} \\ \cmidrule(lr){3-7} \cmidrule(lr){8-11}
$t$ & Model &
\multicolumn{1}{c}{C-10} & \multicolumn{1}{c}{C-20} & \multicolumn{1}{c}{T-10} & \multicolumn{1}{c}{T-20} & \multicolumn{1}{c}{I-100} &
\multicolumn{1}{c}{FC-10} & \multicolumn{1}{c}{FC-20} & \multicolumn{1}{c}{FE-10} & \multicolumn{1}{c}{FE-20}  \\
\midrule

\multirow{2}{*}{$1$}
& HAT & 
$2.7\std{0.8}$ & $12.6\std{1.6}$ & 
$4.0\std{0.3}$ & $0.2\std{0.1}$ &
$25.7\std{1.4}$ & 
$0.3\std{0.0}$ & $6.5\std{0.9}$ &
$27.7\std{2.3}$ & $22.7\std{2.1}$ \\
& \textbf{SPG} & 
$0.1\std{0.0}$ & $0.1\std{0.0}$ &
$0.1\std{0.0}$ & $0.1\std{0.1}$ &
$0.0\std{0.0}$ &
$0.1\std{0.0}$ & $0.2\std{0.0}$ &
$0.1\std{0.0}$ & $0.1\std{0.0}$ \\

\hdashline

\multirow{2}{*}{$T/2$}
& HAT & 
$29.7\std{1.4}$ & $98.3\std{0.6}$ &
$47.7\std{2.1}$ & $29.4\std{2.2}$ &
$99.9\std{0.1}$ &
$3.8\std{0.4}$ & $70.9\std{3.7}$ &
$90.2\std{1.1}$ & $95.6\std{1.4}$ \\
& \textbf{SPG} & 
$0.8\std{0.2}$ & $2.4\std{0.2}$ &
$0.4\std{0.1}$ & $1.4\std{0.2}$ &
$1.6\std{0.2}$ &
$0.8\std{0.1}$ & $1.4\std{0.1}$ &
$0.6\std{0.2}$ & $0.8\std{0.1}$ \\

\hdashline

\multirow{2}{*}{$T$}
& HAT & 
$52.4\std{1.1}$ & $99.7\std{0.3}$ &
$72.3\std{1.6}$ & $50.6\std{2.5}$ &
$99.9\std{0.1}$ & 
$12.1\std{1.3}$ & $84.0\std{1.8}$ &
$98.5\std{0.4}$ & $98.2\std{0.7}$ \\
& \textbf{SPG} & 
$1.6\std{0.2}$ & $4.1\std{0.3}$ &
$0.7\std{0.1}$ & $2.5\std{0.2}$ &
$2.3\std{0.4}$ &
$1.5\std{0.2}$ & $2.2\std{0.3}$ &
$0.9\std{0.4}$ & $1.2\std{0.2}$ \\

\bottomrule
\end{tabular}
}
\end{minipage}

\vspace{2mm}

\begin{minipage}{.85\hsize}
\centering
\subcaption{Results for the 5th fully-connected layer.}
\resizebox{\columnwidth}{!}{
\begin{tabular}{ccrrrrrrrrr}
\toprule
& & \multicolumn{5}{c}{Dissimilar tasks} & \multicolumn{4}{c}{Similar tasks} \\ \cmidrule(lr){3-7} \cmidrule(lr){8-11}
$t$ & Model &
\multicolumn{1}{c}{C-10} & \multicolumn{1}{c}{C-20} & \multicolumn{1}{c}{T-10} & \multicolumn{1}{c}{T-20} & \multicolumn{1}{c}{I-100} &
\multicolumn{1}{c}{FC-10} & \multicolumn{1}{c}{FC-20} & \multicolumn{1}{c}{FE-10} & \multicolumn{1}{c}{FE-20}  \\
\midrule

\multirow{2}{*}{$1$}
& HAT & 
$1.3\std{0.4}$ & $17.9\std{1.0}$ & 
$2.1\std{0.2}$ & $0.1\std{0.0}$ &
$22.0\std{0.5}$ & 
$0.2\std{0.0}$ & $6.0\std{1.0}$ &
$20.8\std{0.5}$ & $16.2\std{1.3}$ \\
& \textbf{SPG} & 
$0.1\std{0.0}$ & $0.2\std{0.0}$ &
$0.1\std{0.0}$ & $0.0\std{0.0}$ &
$0.1\std{0.0}$ &
$0.1\std{0.0}$ & $0.1\std{0.0}$ &
$0.1\std{0.0}$ & $0.1\std{0.0}$ \\

\hdashline

\multirow{2}{*}{$T/2$}
& HAT & 
$17.6\std{1.5}$ & $98.2\std{0.1}$ &
$29.9\std{2.0}$ & $15.4\std{1.2}$ &
$99.8\std{0.0}$ &
$2.1\std{0.2}$ & $61.8\std{1.8}$ &
$84.6\std{0.8}$ & $92.8\std{1.8}$ \\
& \textbf{SPG} & 
$1.0\std{0.1}$ & $3.2\std{0.4}$ &
$0.5\std{0.1}$ & $1.5\std{0.2}$ &
$5.5\std{0.6}$ &
$0.7\std{0.0}$ & $1.2\std{0.2}$ &
$0.7\std{0.3}$ & $1.1\std{0.2}$ \\

\hdashline

\multirow{2}{*}{$T$}
& HAT & 
$34.9\std{2.0}$ & $99.6\std{0.1}$ &
$48.5\std{2.1}$ & $30.6\std{2.0}$ &
$99.9\std{0.1}$ & 
$6.1\std{1.1}$ & $76.4\std{0.6}$ &
$97.4\std{0.3}$ & $97.1\std{1.1}$ \\
& \textbf{SPG} & 
$2.2\std{0.3}$ & $5.8\std{0.7}$ &
$1.1\std{0.2}$ & $3.4\std{0.3}$ &
$8.6\std{0.7}$ &
$1.3\std{0.0}$ & $1.9\std{0.2}$ &
$1.2\std{0.6}$ & $1.8\std{0.3}$ \\

\bottomrule
\end{tabular}
}
\end{minipage}

\end{table}

As we described in \cref{sec:capacity}, SPG blocks much fewer parameters than what HAT does in all cases.
Additionally, we can see from \cref{tab:app_blocking_ratio_layerwise} the significant difference between HAT and SPG in their layer-wise tendency. 
HAT blocks more parameters in earlier layers (e.g., after learning task 5 of C-10, 77.8\% of parameters in the 1st convolution layer are completely blocked while 52.4\% of the ones in the 2nd convolution layer are), which is reasonable given that the earlier layers are supposed to extract basic features and thus changing their parameters without being blocked could easily cause more forgetting than in later layers. 
On the other hand, SPG contrarily tends to completely block more parameters in later layers (e.g., after learning task 5 of C-10, 0.0\% of parameters in the 1st convolution layer are completely blocked while 0.6\% of ones in 2nd convolution layer are). Since SPG computes parameters' importance based on their gradients with regard to the loss through normalization, this result implies that later layers are likely to have more parameters on which some of the tasks highly depend. It can be said that SPG keeps earlier layers alive with less blocking for better basic feature learning (i.e., leading to positive knowledge transfer) while it blocks some specific parameters in later layers that are supposed to be important for previous tasks, which is different from what HAT does.

\newpage

\section{Quantitative Analysis on Cross-Head Importance (CHI)} \label{sec:app_chi}

We analyze in detail how CHI quantitatively contributes to suppressing parameter updates with the following four metrics.

\textbf{(1) Overwrite \underline{\textit{F}}requency at \underline{\textit{each}} task (F-each)}:
How often does the importance from CHI have a larger value than the one from the current task (1 means that it always happens)? It corresponds to cases where $\bm{\gamma}_i^{t,\tau} > \bm{\gamma}_i^{t,t}$ for any $\tau (1 \le \tau < t)$ in \cref{eq:gamma_i_t}. 
\\
\textbf{(2) Overwrite \underline{\textit{G}}ap at \underline{\textit{each}} task (G-each)}:
When the cases of F-each happen, how much is the difference of overwriting on average? It is defined by the average of $\max{(\bm{\gamma}_i^{t,1},\cdots,\bm{\gamma}_i^{t,t-1})} - \bm{\gamma}_i^{t,t}$.
\\
\textbf{(3) Overwrite \underline{\textit{F}}requency in \underline{\textit{total}} (F-total)}:
How often does the importance from CHI actually overwrite the accumulated importance through the maximum operation? It corresponds to cases where $\bm{\gamma}_i^{t,\tau} > \bm{\gamma}_i^{\le t-1}$ for any $\tau (1 \le \tau < t)$ in \cref{eq:gamma_i_le_t}. 
\\
\textbf{(4) Overwrite \underline{\textit{G}}ap in \underline{\textit{total}} (G-total)}:
When the cases of F-total happen, how much is the difference of overwriting on average? It is defined by the average of $\max{(\bm{\gamma}_i^{t,1},\cdots,\bm{\gamma}_i^{t,t-1})} - \bm{\gamma}_i^{\le t-1}$.

The result is presented in \cref{tab:app_ablation_chi}.
We can clearly observe that CHI adds more importance to some parameters (e.g., in C-10, about 15-42\% of parameters constantly update their accumulated importance by the ones from CHI), which is denoted by F-total. Since we introduce CHI to further mitigate forgetting by accumulating more importance, this expectation is consistent with the observed results. Although CHI also overwrites the accumulated importance in similar tasks as frequently as in dissimilar tasks (see F-each and F-total), it happens with a smaller gap overall (see G-each and G-total), which is reasonable as the tasks are similar thus parameters can have similar gradients among different tasks.

\begin{table}[tb]
\centering

\caption{
Quantitative contribution of CHI in learning task $t$. $T$ is the total number of tasks (e.g., $T=10$ for C-10). 
}
\label{tab:app_ablation_chi}

\resizebox{\columnwidth}{!}{
\begin{tabular}{crrrrrrrrrrrr}
\toprule
& \multicolumn{4}{c}{C-10} & \multicolumn{4}{c}{C-20} & \multicolumn{4}{c}{T-10} \\ 
\cmidrule(lr){2-5} \cmidrule(lr){6-9} \cmidrule(lr){10-13}
$t$ & 
\multicolumn{1}{c}{F-each} & \multicolumn{1}{c}{G-each} & \multicolumn{1}{c}{F-total} & \multicolumn{1}{c}{G-total} & 
\multicolumn{1}{c}{F-each} & \multicolumn{1}{c}{G-each} & \multicolumn{1}{c}{F-total} & \multicolumn{1}{c}{G-total} &
\multicolumn{1}{c}{F-each} & \multicolumn{1}{c}{G-each} & \multicolumn{1}{c}{F-total} & \multicolumn{1}{c}{G-total} \\
\midrule

$2$ &
$0.64\std{0.02}$ & $0.15\std{0.02}$ & $0.42\std{0.02}$ & $0.13\std{0.00}$ &
$0.52\std{0.09}$ & $0.10\std{0.03}$ & $0.35\std{0.06}$ & $0.09\std{0.02}$ &
$0.49\std{0.01}$ & $0.13\std{0.00}$ & $0.32\std{0.01}$ & $0.12\std{0.00}$ \\

$T/2$ &
$0.84\std{0.01}$ & $0.31\std{0.02}$ & $0.24\std{0.01}$ & $0.04\std{0.00}$ &
$0.88\std{0.01}$ & $0.33\std{0.02}$ & $0.15\std{0.03}$ & $0.01\std{0.00}$ &
$0.77\std{0.01}$ & $0.28\std{0.01}$ & $0.24\std{0.01}$ & $0.04\std{0.00}$ \\

$T$ &
$0.90\std{0.03}$ & $0.39\std{0.04}$ & $0.15\std{0.01}$ & $0.01\std{0.00}$ &
$0.90\std{0.00}$ & $0.34\std{0.01}$ & $0.05\std{0.00}$ & $0.00\std{0.00}$ &
$0.87\std{0.01}$ & $0.36\std{0.01}$ & $0.13\std{0.00}$ & $0.01\std{0.00}$ \\

\bottomrule
\end{tabular}
}

\resizebox{\columnwidth}{!}{
\begin{tabular}{crrrrrrrrrrrr}
\toprule
& \multicolumn{4}{c}{T-20} & \multicolumn{4}{c}{I-100} & \multicolumn{4}{c}{FC-10} \\ 
\cmidrule(lr){2-5} \cmidrule(lr){6-9} \cmidrule(lr){10-13}
$t$ & 
\multicolumn{1}{c}{F-each} & \multicolumn{1}{c}{G-each} & \multicolumn{1}{c}{F-total} & \multicolumn{1}{c}{G-total} & 
\multicolumn{1}{c}{F-each} & \multicolumn{1}{c}{G-each} & \multicolumn{1}{c}{F-total} & \multicolumn{1}{c}{G-total} &
\multicolumn{1}{c}{F-each} & \multicolumn{1}{c}{G-each} & \multicolumn{1}{c}{F-total} & \multicolumn{1}{c}{G-total} \\
\midrule

$2$ &
$0.52\std{0.01}$ & $0.14\std{0.01}$ & $0.34\std{0.02}$ & $0.12\std{0.01}$ &
$0.47\std{0.01}$ & $0.11\std{0.00}$ & $0.30\std{0.02}$ & $0.11\std{0.01}$ &
$0.53\std{0.04}$ & $0.14\std{0.01}$ & $0.38\std{0.03}$ & $0.12\std{0.00}$ \\

$T/2$ &
$0.88\std{0.02}$ & $0.37\std{0.02}$ & $0.14\std{0.00}$ & $0.01\std{0.00}$ &
$0.91\std{0.00}$ & $0.36\std{0.00}$ & $0.02\std{0.00}$ & $0.00\std{0.00}$ &
$0.83\std{0.06}$ & $0.25\std{0.01}$ & $0.25\std{0.01}$ & $0.04\std{0.00}$ \\

$T$ &
$0.92\std{0.01}$ & $0.41\std{0.02}$ & $0.05\std{0.00}$ & $0.00\std{0.00}$ &
$0.91\std{0.01}$ & $0.37\std{0.00}$ & $0.02\std{0.00}$ & $0.00\std{0.00}$ &
$0.85\std{0.02}$ & $0.28\std{0.01}$ & $0.15\std{0.02}$ & $0.01\std{0.00}$ \\

\bottomrule
\end{tabular}
}

\resizebox{\columnwidth}{!}{
\begin{tabular}{crrrrrrrrrrrr}
\toprule
& \multicolumn{4}{c}{FC-20} & \multicolumn{4}{c}{FE-10} & \multicolumn{4}{c}{FE-20} \\ 
\cmidrule(lr){2-5} \cmidrule(lr){6-9} \cmidrule(lr){10-13}
$t$ & 
\multicolumn{1}{c}{F-each} & \multicolumn{1}{c}{G-each} & \multicolumn{1}{c}{F-total} & \multicolumn{1}{c}{G-total} & 
\multicolumn{1}{c}{F-each} & \multicolumn{1}{c}{G-each} & \multicolumn{1}{c}{F-total} & \multicolumn{1}{c}{G-total} &
\multicolumn{1}{c}{F-each} & \multicolumn{1}{c}{G-each} & \multicolumn{1}{c}{F-total} & \multicolumn{1}{c}{G-total} \\
\midrule

$2$ &
$0.41\std{0.07}$ & $0.10\std{0.01}$ & $0.29\std{0.04}$ & $0.10\std{0.01}$ &
$0.51\std{0.15}$ & $0.08\std{0.02}$ & $0.34\std{0.11}$ & $0.07\std{0.02}$ &
$0.49\std{0.08}$ & $0.09\std{0.01}$ & $0.33\std{0.08}$ & $0.08\std{0.01}$ \\

$T/2$ & 
$0.79\std{0.08}$ & $0.24\std{0.01}$ & $0.12\std{0.01}$ & $0.01\std{0.00}$ &
$0.72\std{0.09}$ & $0.17\std{0.04}$ & $0.25\std{0.07}$ & $0.02\std{0.00}$ &
$0.75\std{0.04}$ & $0.26\std{0.02}$ & $0.13\std{0.01}$ & $0.01\std{0.00}$ \\

$T$ &
$0.84\std{0.06}$ & $0.30\std{0.01}$ & $0.08\std{0.01}$ & $0.00\std{0.00}$ &
$0.86\std{0.02}$ & $0.24\std{0.05}$ & $0.11\std{0.01}$ & $0.01\std{0.00}$ &
$0.89\std{0.00}$ & $0.31\std{0.01}$ & $0.09\std{0.03}$ & $0.00\std{0.00}$ \\

\bottomrule
\end{tabular}
}

\end{table}

\begin{figure}[t]

\centering
\begin{minipage}{.3\hsize}
\centering
\includegraphics[width=\hsize]{images/repr_c10_t_leg.pdf}
\subcaption{Fine-tuning for TinyImageNet after CL for C-10}
\end{minipage}
\hfill
\begin{minipage}{.3\hsize}
\centering
\includegraphics[width=\hsize]{images/repr_i100_c.pdf}
\subcaption{Fine-tuning for CIFAR100 after CL for I-100}
\end{minipage}
\hfill
\begin{minipage}{.3\hsize}
\centering
\includegraphics[width=\hsize]{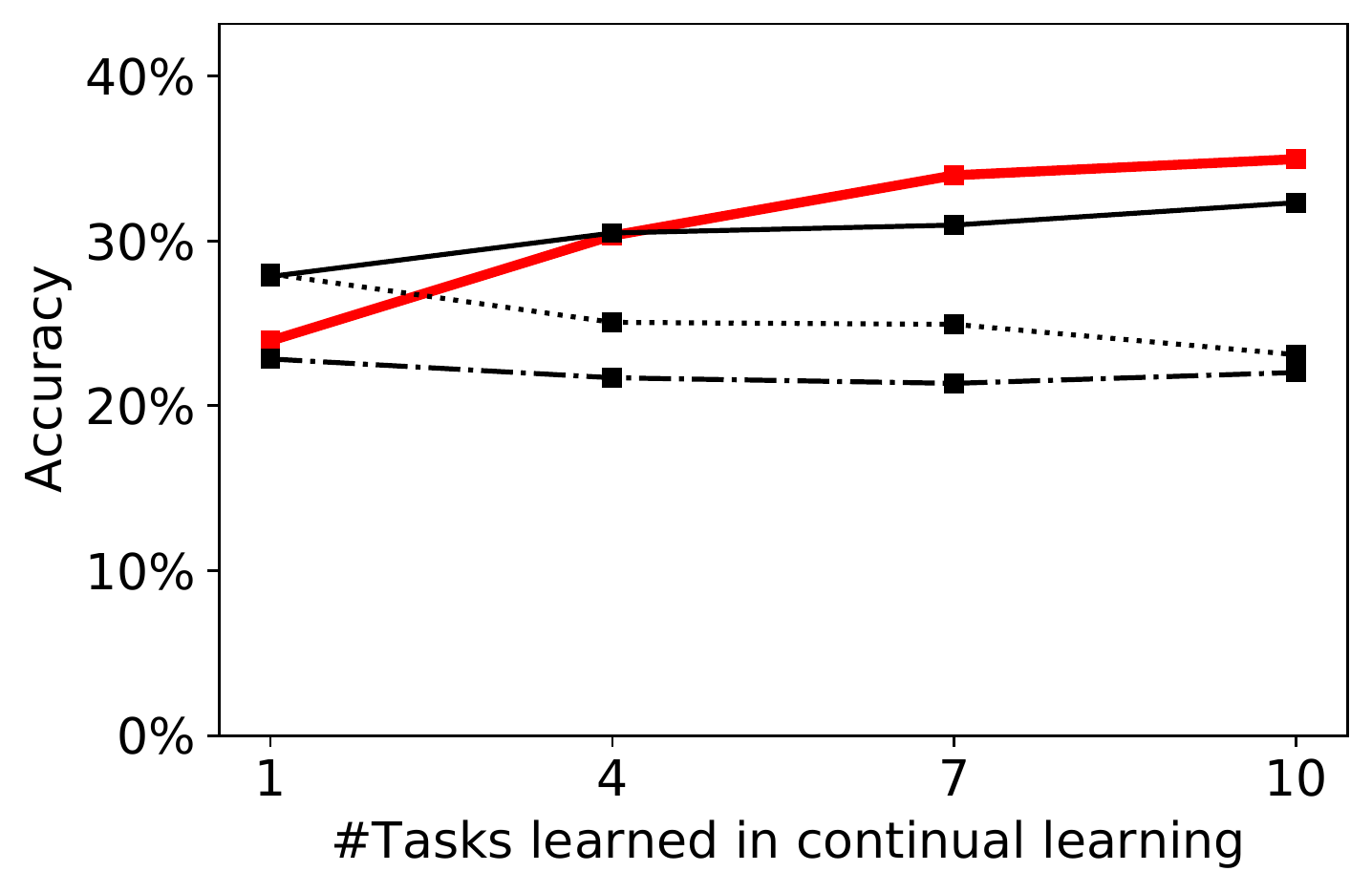}
\subcaption{Fine-tuning for CIFAR100 after CL for T-10}
\end{minipage}

\caption{
The learning of representation through continual learning.
The x-axis means the number of tasks learned in continual learning (CL).
The pair of a CL/non-CL for (a), (b) and (c) is C-10/TinyImageNet, I-100/CIFAR100, and T-10/CIFAR100, respectively.
}
\label{fig:cl_representation_app}

\end{figure}

\section{Representation Learning in Continual Learning} \label{sec:app_representation}

All results (i.e., the pairs of a CL and non-CL dataset are C-10/TinyImageNet, I-100/CIFAR100, and T-10/CIFAR100) are presented in \cref{fig:cl_representation_app}.
We can see that SPG learns better representations in continual learning than baselines in all cases.

\section{Network Size} \label{sec:app_network_size}

The number of learnable parameters of each system is presented in \cref{tab:app_network_size}. Note that all approaches adopt AlexNet as their backbone, and the number of parameters vary depending on their additional structures such as attention mechanisms or sub-modules. It also depends on datasets because each dataset has a different number of tasks and in TIL, each task has a different classification head and the number of units in each classification head depends on the number of classes in each task.
It can be seen that CAT and SupSup need more parameters than SPG and other approaches.

\begin{table}[ht]
\centering
\caption{
The number of learnable parameters of each model. 
``M'' means a million ($1,000,000$).
}
\label{tab:app_network_size}
\begin{tabular}{crrrrrrrrr}
\toprule
& \multicolumn{5}{c}{Dissimilar tasks} & \multicolumn{4}{c}{Similar tasks} \\
\cmidrule(lr){2-6} \cmidrule(lr){7-10}
Model & \multicolumn{1}{c}{C-10} & \multicolumn{1}{c}{C-20} & \multicolumn{1}{c}{T-10} & \multicolumn{1}{c}{T-20} & \multicolumn{1}{c}{I-100} &
\multicolumn{1}{c}{FC-10} & \multicolumn{1}{c}{FC-20} & \multicolumn{1}{c}{FE-10} & \multicolumn{1}{c}{FE-20} \\
\midrule

(MTL) & 
$6.7$M & $6.7$M & 
$6.9$M & $6.9$M &
$8.6$M & 
$6.5$M & $6.6$M &
$7.7$M & $9.0$M \\

(ONE) &
$6.7$M & $6.7$M & 
$6.9$M & $6.9$M &
$8.6$M & 
$6.5$M & $6.6$M &
$7.7$M & $9.0$M \\

\hdashline

NCL &
$6.7$M & $6.7$M & 
$6.9$M & $6.9$M &
$8.6$M & 
$6.5$M & $6.6$M &
$7.7$M & $9.0$M \\

A-GEM &
$6.7$M & $6.7$M & 
$6.9$M & $6.9$M &
$8.6$M & 
$6.5$M & $6.6$M &
$7.7$M & $9.0$M \\

PGN &
$6.7$M & $6.7$M &
$6.7$M & $6.7$M &
$8.3$M & 
$6.0$M & $6.6$M &
$7.5$M & $8.9$M \\

PathNet &
$6.6$M & $6.8$M & 
$6.7$M & $6.6$M &
$8.4$M & %
$6.4$M & $6.4$M &
$7.8$M & $8.7$M \\

HAT &
$6.8$M & $6.8$M &
$7.0$M & $7.0$M &
$9.0$M & 
$6.6$M & $6.7$M &
$7.8$M & $9.1$M \\

CAT &
$39.5$M & $39.7$M &
$40.8$M & $40.9$M &
N/A & 
$38.5$M & $38.9$M &
$46.2$M & $55.1$M \\

SupSup &
$65.2$M & $130.2$M &
$65.4$M & $130.4$M &
$652.0$M &
$65.0$M & $130.1$M &
$65.8$M & $131.7$M \\

UCL &
$6.7$M & $6.7$M &
$6.9$M & $6.9$M &
$8.6$M &
$6.5$M & $6.6$M &
$7.7$M & $9.0$M \\

SI &
$6.7$M & $6.7$M & 
$6.9$M & $6.9$M &
$8.6$M & 
$6.5$M & $6.6$M &
$7.7$M & $9.0$M \\

TAG &
$6.7$M & $6.7$M & 
$6.9$M & $6.9$M &
$8.6$M & 
$6.5$M & $6.6$M &
$7.7$M & $9.0$M \\

WSN &
$6.7$M & $6.7$M &
$6.8$M & $6.8$M &
$8.5$M & 
$6.5$M & $6.5$M &
$7.7$M & $8.9$M \\

EWC &
$6.7$M & $6.7$M & 
$6.9$M & $6.9$M &
$8.6$M & 
$6.5$M & $6.6$M &
$7.7$M & $9.0$M \\

\hdashline

EWC-GI &
$6.7$M & $6.7$M & 
$6.9$M & $6.9$M &
$8.6$M & 
$6.5$M & $6.6$M &
$7.7$M & $9.0$M \\

SPG-FI &
$6.7$M & $6.7$M & 
$6.9$M & $6.9$M &
$8.6$M & 
$6.5$M & $6.6$M &
$7.7$M & $9.0$M \\

\midrule

\textbf{SPG} &
$6.7$M & $6.7$M & 
$6.9$M & $6.9$M &
$8.6$M & 
$6.5$M & $6.6$M &
$7.7$M & $9.0$M \\

\bottomrule
\end{tabular}
\end{table}


\end{document}